\documentclass[fleqn,10pt]{wlscirep}
\usepackage{longtable,array,adjustbox,placeins,float,algorithm,algpseudocode}
\hypersetup{colorlinks=true,allcolors=black}
\definecolor{color1}{RGB}{0,0,0}
\newcounter{papersection}
\renewcommand{\thepapersection}{\arabic{papersection}}

\newcommand{\papersection}[1]{%
  \refstepcounter{papersection}%
  \setcounter{equation}{0}%
  \section*{\thepapersection\ #1}%
}
\newcommand{\figref}[1]{Fig.~\ref{#1}}
\newcommand{\figfullref}[1]{Figure~\ref{#1}}
\newcommand{\tabref}[1]{Table~\ref{#1}}
\newcommand{\eqnref}[1]{Eq.~\eqref{#1}}
\newcommand{\secref}[1]{Section~\ref{#1}}
\newcommand{\secreftext}[2]{\hyperref[#1]{#2}}
\newcommand{\algref}[1]{Algorithm~\ref{#1}}

\title{Earth World Model: Physics-constrained latent dynamics for forecasting and environmental scenario analysis}
\author{%
Huijun Qiao\textsuperscript{1,*},\quad
Ziqiang Zhang\textsuperscript{2,*},\quad
Jinnong Liao\textsuperscript{1,2}
\par\vspace{6pt}
{\Affilfont
\setlength{\baselineskip}{10pt}
\textsuperscript{1}Beijing Latent Vector Technology Co., Ltd.,
Beijing 100091, China
\par\vspace{-2pt}
\textsuperscript{2}School of Mechanical and Energy Engineering,
Beijing University of Technology, Beijing 100124, China
\par\vspace{2pt}
\textsuperscript{*}Co-corresponding authors:
\href{mailto:jooojoe@gmail.com}{jooojoe@gmail.com};\quad
\href{mailto:zhangziqiang@bjut.edu.cn}{zhangziqiang@bjut.edu.cn}.
}
}

\begin{abstract}
World models have been widely applied to multi-agent planning and forecasting, but exogenous environmental changes have yet to be incorporated as replaceable drivers within a unified state-evolution process. Consequently, existing models offer limited means to assess, from an Earth-centred perspective, whether environmental perturbations propagate through physical systems with faithful dynamics. Here we propose the Earth World Model (EWM), an Earth-centred forecasting framework. EWM ingests heterogeneous multi-source observations and compresses them into a reusable low-dimensional latent vector that encodes the system's history and current state. Given explicit exogenous drivers, a shared transition operator recursively updates this latent state, and physics-constrained decoding then produces multi-step forecasts. Substituting the exogenous drivers further enables environmental scenario analysis and tests the fidelity of the resulting dynamic responses.
The framework was evaluated for power forecasting and environmental-perturbation response at six photovoltaic sites and one wind-power site. Across four photovoltaic sites and three forecast horizons, three independent training runs yielded mean $R^2$ values from 0.8578 to 0.9095; performance remained stable across sites and horizons. In a separate six-site decoder ablation, the mean net per-unit response $\rho$ under the prescribed irradiance intervention increased from 0.180 with the multilayer perceptron (MLP) decoder to 0.922 with the analytic physical decoder, close to the corresponding site- and horizon-specific analytic references. The wind model achieved an injection elasticity of $0.9990 \pm 0.0006$, while its pre-intervention outputs remained unchanged, indicating accurate transmission of perturbation amplitude and preservation of its temporal boundary. These experiments show that forecast fit and intervention response require separate evaluation, and that the evaluated EWM configurations preserved response amplitude and temporal boundaries under the prescribed interventions.

\end{abstract}
\begin{document}
\maketitle
\thispagestyle{empty}
\raggedbottom
\renewcommand{\topfraction}{0.95}
\renewcommand{\textfraction}{0.05}
\renewcommand{\floatpagefraction}{0.9}

\FloatBarrier

\papersection{Introduction}\label{sec:introduction}

External Earth-system forcing prompts responses from human activities, ecological processes, and engineered infrastructure. Solar radiation, atmospheric motion, wind, cloud, precipitation and temperature act on evolving surface states that shape energy production, crop growth, infrastructure exposure and environmental risk \cite{reichstein2019earth,Irrgang2021neuralEarth}. An Earth-centred world model must therefore represent both how an issued environmental driver changes a surface system and how that change reaches a decision-relevant output. This requirement extends beyond fitting a task-specific target because the same model must support forecasting, response analysis and specified what-if queries.

Existing world models approach these challenges through three main technical routes. First, agent-centred world models \cite{ha2018worldModels,hafner2020dreamer} learn latent regularities \cite{bruce2024genie} and simulate the possible consequences of actions \cite{schrittwieser2020muzero} to support agent planning \cite{hafner2025dreamer}. Second, weather-forecasting models \cite{chen2024fuxi,price2025gencast} learn representations of large-scale atmospheric and land-surface systems \cite{kochkov2024neuralgcm,allen2025aardvark}, improving medium-range forecast accuracy \cite{lam2023graphcast,bi2023pangu} and cross-variable transfer \cite{bodnar2025aurora}. Third, physics-informed and physics-constrained approaches \cite{karniadakis2021physics,liu2025physicsStructure} incorporate physical laws through equation residuals \cite{raissi2019pinn,chen2021scarceData}, analytical relations \cite{beucler2021analyticConstraints} or differentiable physical operators \cite{li2024pino,azizzadenesheli2024neuralOperators}. Agent-centred world models, however, do not treat environmental forcing as an external control variable. Existing weather-forecasting models generally use future meteorological variables as conditioning inputs rather than as replaceable exogenous drivers within a unified state-transition process. Physics-informed approaches, meanwhile, often introduce physical constraints only as penalty terms in the loss function, without directly constraining the dynamic evolution of system states and the output-decoding process. Existing methods therefore have difficulty representing, within a unified framework, the dynamic evolution and physical responses of Earth-surface systems under external environmental forcing. This limitation reduces forecasting reliability and scenario-analysis capability under complex environmental changes, hindering the trustworthy prediction and scientific decision-making required in energy production, environmental and ecological management, and engineering infrastructure.

To address these limitations, we propose the Earth World Model (EWM), defined as the tuple \(\mathcal{M}=(Z,T,C,D)\). At forecast issue time, a boundary representation condensation (BRC) encoder draws on the multi-source observations and historical environmental drivers available at that time, compressing them into a low-dimensional latent-state vector in \(Z\) that encodes the system's history and current state. The shared state-transition operator \(T\) then recursively carries this latent state forward, using the explicit future exogenous drivers in \(C\). A physical decoder \(D\) maps each evolved latent state and its corresponding environmental driver to a decision-relevant output. By combining recursive rollout with replaceable exogenous drivers, EWM unifies multi-step forecasting and environmental scenario simulation within a single computation. Keeping the transition operator and decoder fixed across baseline and query runs isolates the model response to the declared driver replacement under matched computational conditions. The resulting quantities characterize response timing, amplitude and input support within the evaluated model interface. Physical-causality claims should be supported by independent evidence from the corresponding physical system.

We evaluate EWM using photovoltaic and wind-power forecasting as representative applications. Both systems are strongly driven by meteorological conditions \cite{verdone2025forecastReview}: their power outputs depend not only on environmental factors such as solar irradiance, temperature \cite{skoplaki2009temperature,zhang2026crossUnet} and wind speed \cite{lydia2014powerCurve,liu2024bayesianWind}, but also on system operating states and nonlinear energy-conversion relationships \cite{gao2025physicsWind}. Conventional data-driven forecasting methods \cite{meng2024metaWind,meng2025limitedData} focus primarily on the mapping from inputs to power output \cite{wang2025multimodalPV} and therefore have difficulty explicitly representing the state evolution and physical responses induced by changes in environmental drivers. We address this issue through comparative experiments and ablation analyses spanning three dimensions: power-forecasting performance, physical decoding and responses to environmental perturbations. These experiments systematically assess the multi-step forecasting, physical consistency and scenario-analysis capabilities of EWM. The results show that EWM maintains good power-forecasting performance while reducing attenuation in the response to environmental drivers and preserving the temporal consistency of perturbation propagation, supporting its effectiveness for renewable-power forecasting and environmental scenario analysis.

\FloatBarrier

\papersection{EWM Definition and Overall Framework}\label{sec:ewm}

An Earth World Model represents a coupled Earth--surface process with a latent state, an explicit external drive, a transition operator and a decision-relevant decoder. We write the model as

\begin{equation}\label{eq:2.1}\begin{adjustbox}{max width=.92\linewidth}$\displaystyle \mathcal{M}=({Z},{T},{C},{D})$\end{adjustbox}\end{equation}

where \(Z\) is the low-dimensional latent-state space used to represent the system history and current state; \(C\) is the space of explicit exogenous drivers describing the external environmental conditions that influence system evolution; \(T:Z\times C\rightarrow Z\) is the transition operator that dynamically couples the exogenous drivers to the latent state and advances the system state recursively through time; and \(D\) is the physical decoder that maps an evolved latent state and the corresponding exogenous driver, according to the relevant physical relationships, to a target output with a clear physical meaning. Together, these four elements define a unified dynamical modelling framework that links external environmental forcing, system-state evolution and physically meaningful outputs.

\begin{figure}[!t]\centering
\includegraphics[width=\linewidth,pagebox=cropbox]{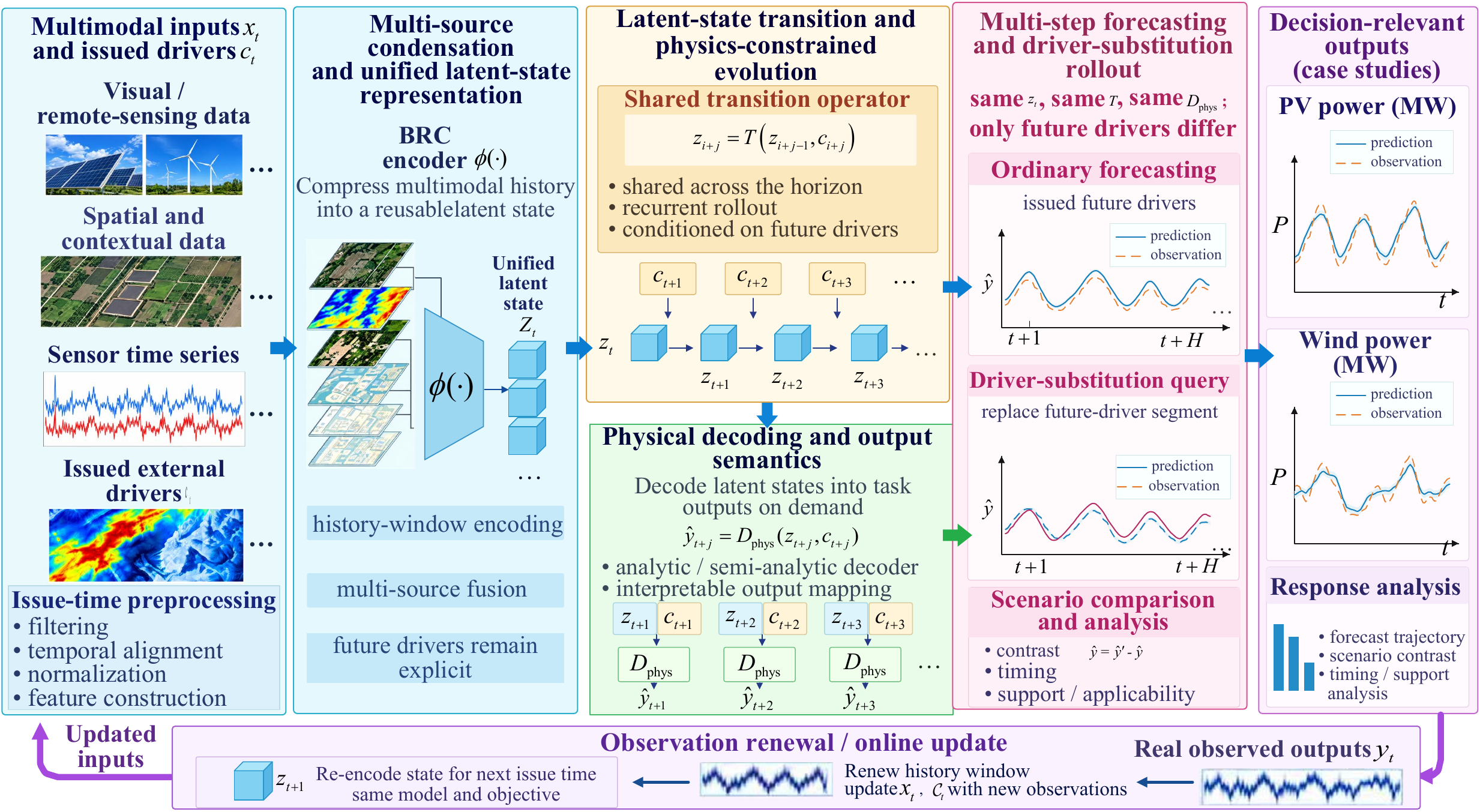}
\caption{EWM framework. Multimodal observations and issued environmental drivers are condensed into a reusable low-dimensional latent state, advanced recurrently by a shared driver-conditioned transition, decoded on demand through physical relations, and used for ordinary forecasting or driver-substitution rollout. New observations renew the history window and encoded state at the next issue time.}\label{fig:ewm}
\end{figure}

\figfullref{fig:ewm} presents the framework of the Earth World Model (EWM) and its four modules: multi-source information fusion and latent-state representation, latent-state transition and physical evolution, physics-constrained decoding, and multi-step forecasting and environmental scenario analysis. Encoding begins by jointly processing multi-source observations and historical environmental drivers, which are compressed into a reusable low-dimensional latent-state vector that gives a unified description of the system's history and current state. Future environmental variables then serve as explicit exogenous drivers, and the shared state-transition operator recursively updates the latent state, generating a dynamic trajectory that represents the system response to environmental change. Domain-specific physical relationships and operating constraints are subsequently incorporated into the physical decoder, which converts the evolved latent states into target outputs with direct physical meaning and thereby produces multi-step forecasts. Environmental scenarios are generated by replacing selected exogenous drivers while holding the initial state and model parameters fixed. The resulting state and output trajectories allow the temporal consistency, response amplitude and physical plausibility of perturbation propagation to be evaluated. When new observations become available, the model re-encodes the updated history to renew the latent state for subsequent forecasting and scenario analysis. EWM thus establishes a unified computational mechanism that links multi-source information representation, dynamic state evolution, physical outputs and environmental scenario analysis.

The two structural constraints below define the computational boundary between EWM and a direct sequence-to-sequence predictor. First, rollout autonomy requires one transition operator to be iterated along the full horizon:

\begin{equation}\label{eq:2.2}\begin{adjustbox}{max width=.92\linewidth}$\displaystyle {z}_{t+j}=T({z}_{t+j-1},{c}_{t+j}),\qquad {\hat y}_{t+j}=D({z}_{t+j},{c}_{t+j}),\qquad j=1,\ldots,H.$\end{adjustbox}\end{equation}

Here $H$ is the forecast horizon and $j$ indexes the recurrent steps. The model does not replace the state trajectory with $H$ independent output heads and does not feed future ground-truth observations into the rollout. History is encoded once; the same transition parameters are reused at each step, so the forecast is generated by a single latent trajectory driven by the issued sequence.

Second, intervention injectability requires a query trajectory to start from the same encoded state and to reuse the same forward path after only the declared external drive has been replaced:

\begin{equation}\label{eq:2.3}\begin{adjustbox}{max width=.92\linewidth}$\displaystyle {z}^{\prime}_{t}={z}_{t},\qquad {z}^{\prime}_{t+j}=T({z}^{\prime}_{t+j-1},{c}^{\prime}_{t+j}),\qquad {\hat y}^{\prime}_{t+j}=D({z}^{\prime}_{t+j},{c}^{\prime}_{t+j}).$\end{adjustbox}\end{equation}

The queried drive $c^{\prime}$ is identical to $c$ except for the declared variable, interval, and level. With deterministic inference and the same handling of derived features, a neutral substitution ($c^{\prime}=c$) is expected to recover the baseline trajectory. If the earliest altered step is $a$, then outputs at $j<a$ serve as the pre-intervention temporal-locality check. These conditions set EWM apart from a conventional sequence-to-sequence forecaster in that they specify both how the horizon is generated and how an external driver enters that path. These tests capture model-level identity and temporal locality; conclusions about physical causality require independent evidence from the physical system.

\papersection{Algorithm Design}\label{sec:algorithm}

\subsection*{3.A Multi-source Condensation and Unified Latent-state Representation}

Boundary representation condensation (BRC) encodes observations and issued historical drivers into a latent state after aligning the information available at a given issue time. It combines heterogeneous inputs within each domain. The PV and wind implementations use separate encoder and transition weights. Let $x$ denote the observation history, $c$ the external-driver sequence, $L$ the history-window length, $n_x$ the number of observation channels, and $A$ the operation that performs availability filtering, temporal alignment, normalization and feature construction. The information available at issue time $t$ is

\begin{equation}\label{eq:3.1}\begin{adjustbox}{max width=.92\linewidth}$\displaystyle {\mathcal I}_{t}=\{r:a(r)\leq t\},\qquad {h}_{t}=A({\mathcal I}_{t})$\end{adjustbox}\end{equation}

where $r$ is an input record and $a(r)$ is its availability time. The initial latent state is

\begin{equation}\label{eq:3.2}\begin{adjustbox}{max width=.92\linewidth}$\displaystyle {z}_{t}=\Phi({h}_{t})=\Phi\left({x}_{t-L+1:t},{c}_{t-L+1:t}\right)$\end{adjustbox}\end{equation}

where $\Phi$ is the BRC encoder and $d_z$ is the latent dimension. A time-by-channel history is divided into local patches, represented jointly across channels and time, pooled and projected to $z_t\in\mathbb{R}^{d_z}$. The BRC encoder uses Patch for local tokenization, Pool for aggregation and Proj for projection. Future drivers $c_{t+1:t+H}$ are excluded from the initial-state encoding and enter the transition only after $z_t$ has been formed. At inference, normalization parameters fitted on the training set remain fixed. In a future-driver substitution query, features derived from historical observations retain their baseline values, whereas features derived from the queried control are recomputed from the substituted value.

This separation assigns a specific computational role to the latent representation. It stores the aligned history once and provides the common initial condition for ordinary forecasting and matched future-driver queries. The decoder is called on demand and maps the evolved latent state and current drive directly to the required output.

\subsection*{3.B Latent-State Transition and Physics-Constrained Evolution}

At forecast step $j$, the transition operator advances the latent state under the current external drive. Its domain-specific implementation determines which state components evolve and which retain information encoded at issue time:

\begin{equation}\label{eq:3.3}{z}_{t+j}=T({z}_{t+j-1},{c}_{t+j}),\qquad j=1,\ldots,H.\end{equation}

The transition parameters are shared across the horizon. In the gated recurrent unit (GRU) implementation, $\operatorname{ctrlproj}$ maps the current drive to the recurrent input and $\operatorname{GRUCell}$ updates the dynamic state. A history-only static component, when present, is carried forward unchanged by $T$; the wind partition is specified in \secreftext{sec:wind-decoder}{Section~4.3.2}. The model encodes history once and recurrently updates its dynamic state using transition parameters shared across forecast steps. At inference, each state is computed from the preceding state and the corresponding issued driver. The same recurrence is used for ordinary forecasting and driver-substitution queries.

The representation objective constrains the information carried by the predicted state without requiring reconstruction of every raw observation. Let $P_{\theta_{\mathrm{repr}}}$ be the representation predictor, $z_{\mathrm{tar}}$ a target representation available during training, and $\operatorname{Dist}$ the discrepancy implemented for the selected domain:

\begin{equation}\label{eq:3.4} {\hat z}_{\mathrm{tar}}=P_{\theta_{\mathrm{repr}}}({z}_{t},{c}_{t+1:t+H}),\qquad {L}_{\mathrm{repr}}=\mathrm{Dist}({\hat z}_{\mathrm{tar}},{z}_{\mathrm{tar}}).\end{equation}

Target representations are used only for training supervision; inference uses the encoded history and issued future drivers. In \eqnref{eq:3.4}, $\theta_{\mathrm{repr}}$ is the parameter subset of the representation predictor. The generic domain objective is written as

\begin{equation}\label{eq:3.5}\begin{adjustbox}{max width=.92\linewidth}$\displaystyle {L}_{d}={L}_{\mathrm{task},d}+\lambda_{\mathrm{repr}}{L}_{\mathrm{repr},d}+\lambda_{\mathrm{phys}}{L}_{\mathrm{phys},d},\qquad d\in\mathcal D.$\end{adjustbox}\end{equation}

where $d\in\mathcal D$ indexes an application domain; $L_{\mathrm{task},d}$ is the output-fitting objective, $L_{\mathrm{repr},d}$ is the optional representation objective and $L_{\mathrm{phys},d}$ collects domain-appropriate physical-response constraints. The non-negative weights $\lambda_{\mathrm{repr}}$ and $\lambda_{\mathrm{phys}}$ set the contributions of the latter two terms. Night-time clamping and operating gates are structural rules rather than additional loss terms. The PV and wind instantiations, including the reported parameter configurations, are specified in \secref{sec:experiments} and \tabref{tab:domain}.

Intervention injectability is the second structural mechanism: a specified change in an external driver enters the shared state-transition process through an explicit input. For a query specification $q$, the operator $\operatorname{Replace}(c,q)$ produces $c^{\prime}$ by changing only the declared drive variable, time interval and level. The baseline and query paths share $z_t$, $T$ and $D$; only the drive sequence differs. The difference between these matched paths gives the model response to the specified forward-input change. A neutral replacement leaves the trajectory unchanged. For a non-neutral replacement, response timing, amplitude and support are evaluated separately.

EWM imposes physical constraints through both its structure and its training objective. The transition determines how external drivers affect the latent state; the decoder maps that state to the output. Ordinary forecasts and driver-substitution queries share these operators and the resulting dynamical model.

In \algref{alg:1}, $d$ identifies the domain, $e$ the training epoch and $j$ the forecast step. Training runs for at most $E_{\max}$ epochs, and the forecast horizon is $H$. The training and validation sets are $\mathcal D_{\mathrm{train}}$ and $\mathcal D_{\mathrm{val}}$, respectively; $\mathcal B$ denotes a mini-batch. The operators $\Phi_d$, $T_d$ and $D_d$ are the domain-specific encoder, transition and decoder. The vector $y_{t+1:t+H}$ contains observed targets and $\hat y_{t+1:t+H}$ the corresponding predictions. $\Theta$ denotes all trainable parameters and contains $\theta_{\mathrm{repr}}$ and the decoder parameters $\psi$ when those components are used. $\nabla_\Theta L_d$ is the loss gradient, $\Theta_e$ is the checkpoint after epoch $e$, and $\ell_e$ is its validation loss. In $\{\ell_k,\Theta_k\}_{k\le e}$, $k$ indexes visited checkpoints; $s$ is the Boolean stopping flag returned with the selected parameters $\Theta^*$. $\operatorname{OptimizerStep}$ performs one parameter update, whereas $\operatorname{Select}_d$ applies the domain-specific validation and stopping rule stated in \secreftext{sec:case-study}{Section~4.1}.

Here $\operatorname{InitializeParameters}_d$ denotes either initialization from a specified pretrained checkpoint or random initialization for training from scratch, as stated for each experiment in \secreftext{sec:case-study}{Section~4.1}. $\operatorname{ValLoss}_{\mathcal D_{\mathrm{val}}}$ evaluates the domain objective on the validation split. TaskLoss, RepresentationLoss and PhysicsLoss implement the three terms in \eqnref{eq:3.5}; an unused optional term contributes zero. In \tabref{tab:domain}, JEPA denotes joint-embedding predictive architecture and MSE denotes mean squared error.

\begin{algorithm}[!t]
\hrule height 0.8pt
\vspace{0.25em}
\refstepcounter{algorithm}\label{alg:1}
\noindent\textbf{Algorithm~\thealgorithm:} Domain-specific latent-state transition training\par
\vspace{0.25em}
\hrule height 0.4pt
\vspace{0.4em}
\noindent\textbf{Input:} domain $d$, training/validation splits, issued drivers, target supervision and optimizer.\par
\noindent\textbf{Output:} selected parameters $\Theta^*$.
\renewcommand{\AlgItem}{\refstepcounter{algline}\item[\scriptsize\arabic{algline}:]}
\begin{algorithmic}[1]
\State $\Theta\gets\operatorname{InitializeParameters}_d$
\For{$e=1,\ldots,E_{\max}$}
\For{each mini-batch $\mathcal B$ in $\mathcal D_{\mathrm{train}}$}
\State $h_t\gets A(\mathcal I_t),\quad z_t\gets\Phi_d(h_t)$
\For{$j=1,\ldots,H$}
\State $z_{t+j}\gets T_d(z_{t+j-1},c_{t+j})$
\State $\hat y_{t+j}\gets D_d(z_{t+j};c_{t+j})$
\EndFor
\State $L_{\mathrm{task},d}\gets\operatorname{TaskLoss}_d(\hat y_{t+1:t+H},y_{t+1:t+H})$
\State $L_{\mathrm{repr},d}\gets\operatorname{RepresentationLoss}_d(z_t,c_{t+1:t+H},z_{\mathrm{tar}})$
\State $L_{\mathrm{phys},d}\gets\operatorname{PhysicsLoss}_d(\hat y_{t+1:t+H},c_{t+1:t+H},z_{t+1:t+H})$
\State $L_d\gets L_{\mathrm{task},d}+\lambda_{\mathrm{repr}}L_{\mathrm{repr},d}+\lambda_{\mathrm{phys}}L_{\mathrm{phys},d}$
\State $\Theta\gets\operatorname{OptimizerStep}(\Theta,\nabla_\Theta L_d)$
\EndFor
\State $\Theta_e\gets\Theta;\quad\ell_e\gets\operatorname{ValLoss}_{\mathcal D_{\mathrm{val}}}(\Theta_e)$
\State $(\Theta^*,s)\gets\operatorname{Select}_d(\{\ell_k,\Theta_k\}_{k\leq e})$
\If{$s=\mathrm{true}$}
\State\textbf{break}
\EndIf
\EndFor
\State\Return $\Theta^*$
\end{algorithmic}
\vspace{0.2em}
{\footnotesize\noindent\textit{Note:} The domain-specific TaskLoss, RepresentationLoss and PhysicsLoss are instantiated only in the case-study section. The recursion uses one transition operator across all steps; target representations are training-only inputs.\par}
\vspace{0.2em}
\hrule height 0.8pt
\end{algorithm}

\subsection*{3.C Physical Decoding}

The physical decoder connects latent dynamics to the quantity that a user can observe or act on. The encoder compresses a high-dimensional history input into a low-dimensional state, and the transition operator stores the evolving consequences of future Earth-system forcing in that state. Because the latent coordinates need not have the units or semantics of a task output, a separate decoder is required to recover power, energy or another decision-relevant variable without reconstructing the full input history.

For forecast step $j$, the decoder is

\begin{equation}\label{eq:3.6}\begin{adjustbox}{max width=.92\linewidth}$\displaystyle {\hat y}_{t+j}=D_{\mathrm{phys}}({z}_{t+j},{c}_{t+j};\psi)=\mathcal P_d\!\left({c}_{t+j},\alpha_{\psi}({z}_{t+j})\right),\qquad j=1,\ldots,H.$\end{adjustbox}\end{equation}

Here ${z}_{t+j}$ is the evolved latent state, ${c}_{t+j}$ is the current external drive, $\mathcal P_d$ is the domain-specific physical mapping and $\alpha_{\psi}$ is a restricted learned correction parameterized by $\psi\subset\Theta$. Decoding is performed on demand. The decoded output does not replace the recurrent state and is not fed back as future ground truth, so state evolution remains governed by the shared transition operator.

The learned correction represents unresolved site- and state-dependent effects that are not described explicitly by the physical mapping. In the PV case, irradiance is converted through cell temperature and temperature derating to power, while a learned factor represents the combined effect of slowly varying site losses. In the wind case, the latent state corrects effective wind speed before an aerodynamic conversion applies the cubic speed relation, rated capacity, and cut-in and cut-out limits. The decoder ablations and response diagnostics in \secreftext{sec:experiments}{Section~4} test whether the correction preserves the response to a substituted driver.

The modular separation permits the role of each component to be tested. The transition determines how an external drive changes the latent trajectory, whereas the decoder determines how that trajectory becomes a physical output. A flexible black-box head can improve fit by absorbing driver error while also attenuating the response to a substituted drive. The decoder ablations in \secreftext{sec:experiments}{Section~4} therefore report forecast fit and response fidelity separately and identify whether an error originates in the drive-transmission path or the final output conversion.

\subsection*{3.D Multi-step Forecasting and Driver-substitution Rollout}

Multi-step forecasting starts from \eqnref{eq:3.2}, successively applies \eqnref{eq:3.3}, and decodes each evolved state with \eqnref{eq:3.6}. The same transition parameters are reused at all steps and future ground-truth outputs are not fed back as state inputs. A driver-substitution query $q$ specifies the queried variable, operation, time interval and replacement level. The supported operations are those implemented for the selected case, including scaling or setting a future-driver segment. The query starts from the same $z_t$ and replaces only that declared segment:

\begin{equation}\label{eq:3.7}\begin{adjustbox}{max width=.92\linewidth}$\displaystyle {z}^{\prime}_{t}={z}_{t},\qquad {z}^{\prime}_{t+j}=T({z}^{\prime}_{t+j-1},{c}^{\prime}_{t+j}),\qquad {\Delta\hat y}_{t+j}={\hat y}^{\prime}_{t+j}-{\hat y}_{t+j}.$\end{adjustbox}\end{equation}

Let $W\subseteq\{1,\ldots,H\}$ be an integration window and let $\Delta t$ be the time step in hours. The corresponding energy summary is

\begin{equation}\label{eq:3.8}\begin{adjustbox}{max width=.92\linewidth}$\displaystyle {E}_{W}=\sum_{j\in W}{\hat y}_{t+j}\,\Delta t.$\end{adjustbox}\end{equation}

If a query changes the driver beginning at step $a$, steps before $a$ define the pre-intervention temporal check. In \algref{alg:2}, $q=\varnothing$ denotes an ordinary forecast without a scenario query; $\operatorname{EvalMode}$ fixes $\Theta^*$ and disables training-time stochastic operations. The vectors $\hat y$ and $\hat y^{\prime}$ are the baseline and substituted-driver forecasts, $\Delta\hat y=\hat y^{\prime}-\hat y$ is their pointwise contrast, and $E_W$ and $E_W^{\prime}$ are the corresponding energy summaries over $W$. With power in MW and $\Delta t$ in hours, energy is in MWh. Relative energy changes require a valid non-zero baseline; the archived material does not establish a common numerical exclusion threshold. $\operatorname{SupportCheck}$ and $\operatorname{Applicability}$ check input support, the pre-window boundary, denominator validity and output limits for each case. These checks are case-specific; their validity as universal tests has not been independently established. $\operatorname{Replace}$ validates the query, recomputes control-derived features and retains history-derived features. $\operatorname{FirstChangedStep}$ then compares the model-input drive sequences over the forecast horizon and returns the earliest changed step $a$.

If the sequences are identical, $a=H+1$. \algref{alg:2} then reuses the baseline forecast, so $\Delta\hat y=0$ and $E_W^{\prime}=E_W$, and still performs the support and applicability checks. Unit scaling, equal-value replacement and a valid query outside the forecast horizon all follow this rule. Every forecast step belongs to the unchanged pre-window; invalid variables or operations are rejected. When a change is present, $\operatorname{Rollout}_d$ recursively applies the transition and decoder from $z_0$. The check results are recorded as $b_{\mathrm{sup}}$ and $b_{\mathrm{app}}$, with $b$ denoting the pair. The next issue time, after new observations become available, is $t^+$.

\begin{algorithm}[!t]
\hrule height 0.8pt
\vspace{0.25em}
\refstepcounter{algorithm}\label{alg:2}
\noindent\textbf{Algorithm~\thealgorithm:} EWM forecasting, driver substitution and observation renewal\par
\vspace{0.25em}
\hrule height 0.4pt
\vspace{0.4em}
\noindent\textbf{Input:} domain $d$, trained parameters $\Theta^*$, issue-time information $\mathcal I_t$, issued drivers $c$, horizon $H$, integration window $W$ and optional scenario $q$.\par
\noindent\textbf{Output:} forecast, scenario contrast and applicability flags.
\renewcommand{\AlgItem}{\refstepcounter{algline}\item[\scriptsize\arabic{algline}:]}
\begin{algorithmic}[1]
\Function{Rollout$_d$}{$z_0,c_{1:H};\Theta$}
\For{$j=1,\ldots,H$}
\State $z_j\gets T_d(z_{j-1},c_j);\quad \hat y_j\gets D_d(z_j;c_j)$
\EndFor
\State\Return $\hat y_{1:H}$
\EndFunction
\State $\Theta\gets\operatorname{EvalMode}(\Theta^*);\quad z_t\gets\Phi_d(A(\mathcal I_t))$
\State $\hat y\gets\operatorname{Rollout}_d(z_t,c_{t+1:t+H};\Theta)$
\State $E_W\gets\sum_{j\in W}\hat y_{t+j}\Delta t$
\If{$q=\varnothing$}
\State\Return $(\hat y,E_W)$
\EndIf
\State $c^\prime\gets\operatorname{Replace}(c,q);\quad a\gets\operatorname{FirstChangedStep}(c,c^\prime);\quad z_t^\prime\gets z_t$
\If{$a=H+1$}
\State $\hat y^\prime\gets\hat y$
\Else
\State $\hat y^\prime\gets\operatorname{Rollout}_d(z_t^\prime,c^\prime_{t+1:t+H};\Theta)$
\EndIf
\State $\Delta\hat y\gets\hat y^\prime-\hat y;\quad E_W^\prime\gets\sum_{j\in W}\hat y^\prime_{t+j}\Delta t$
\State $b_{\mathrm{sup}}\gets\operatorname{SupportCheck}(c^\prime;\mathcal D_{\mathrm{train}})$
\State $b_{\mathrm{app}}\gets\operatorname{Applicability}(\hat y,\hat y^\prime,c,c^\prime,a,W)$
\State $b\gets(b_{\mathrm{sup}},b_{\mathrm{app}})$
\State\Return $(\hat y,\hat y^\prime,\Delta\hat y,E_W,E_W^\prime,b)$
\Procedure{OnNewObservation}{$\mathcal I_{t^+}$}
\State $h_{t^+}\gets A(\mathcal I_{t^+});\quad z_{t^+}\gets\Phi_d(h_{t^+})$
\State\Return $z_{t^+}$
\EndProcedure
\end{algorithmic}
\vspace{0.2em}
{\footnotesize\noindent\textit{Note:} Here $q$ is an intervention specification; $\operatorname{Replace}$ validates the query and updates the declared drive and its derived features; $a$ is the first changed step, or $H+1$ if no step changes; and $D_d$ includes domain operating limits. $b_{\mathrm{sup}}$ and $b_{\mathrm{app}}$ are the support and applicability flags. Every rollout uses matched weights and deterministic evaluation. OnNewObservation updates the encoded state without updating network parameters.\par}
\vspace{0.2em}
\hrule height 0.8pt
\end{algorithm}

\papersection{Experimental results and discussion}\label{sec:experiments}

\subsection*{4.1 Case-study problem definitions, data and evaluation measures}\phantomsection\label{sec:case-study}

PV and wind-power forecasting estimates plant active power over a future horizon from recent observations and meteorological drivers sampled at 15-min and 10-min intervals, respectively. PV conversion depends on irradiance and thermal conditions \cite{skoplaki2009temperature}, whereas wind conversion follows the turbine power curve \cite{lydia2014powerCurve}. \tabref{tab:inputs} summarizes the shared input roles and domain-specific channels. The observation history ${x}_{t-L+1:t}\in\mathbb{R}^{L\times n_x}$ includes measured power and local environmental variables. Driver inputs comprise historical drivers ${c}_{t-L+1:t}$ for state estimation and future drivers ${c}_{t+1:t+H}$ for rollout. PV uses numerical weather prediction (NWP) inputs and cyclic local-time features; its derived irradiance channels are

\begin{equation}\label{eq:4.1}\begin{adjustbox}{max width=.92\linewidth}$\displaystyle {k}_{t\mathrm{,h}}\mathrm{=}\frac{G}{{G}^{\mathrm{cs}}}\mathrm{,}\quad {k}_{t\mathrm{,f}}\mathrm{=}\frac{{G}^{\mathrm{NWP}}}{{G}^{\mathrm{cs}}}\mathrm{,}$\end{adjustbox}\end{equation}
\begin{equation}\label{eq:4.2}\begin{adjustbox}{max width=.92\linewidth}$\displaystyle {\mathrm{bias}}_{24}\left( t\right)\mathrm{=}\frac{1}{{N}_{\mathrm{b}}}\sum_{i\mathrm{=}0}^{{N}_{\mathrm{b}}\mathrm{-}1} \left( {G}_{t\mathrm{-}H\mathrm{-}i}^{\mathrm{NWP}}\mathrm{-}{G}_{t\mathrm{-}H\mathrm{-}i}\right)\mathrm{,}$\end{adjustbox}\end{equation}

where $G$, $G^{\mathrm{NWP}}$ and $G^{\mathrm{cs}}$ denote measured, NWP and clear-sky irradiance. Accordingly, $k_{t,\mathrm h}$ and $k_{t,\mathrm f}$ are the measured and forecast clear-sky indices. Clear-sky ratios are evaluated only when $G^{\mathrm{cs}}>0$. In \eqnref{eq:4.2}, $\mathrm{bias}_{24}(t)$ is the trailing mean NWP-minus-observation irradiance bias, $N_b=96$ is the averaging length (24 h at 15-min resolution), and the horizon shift restricts residuals to timestamps no later than $t-H$. The PV model therefore consumes 18 history channels and 13 future-control channels; the 13 raw normalized columns are a separate preprocessing count, and the three derived channels are appended after normalization. During a future-irradiance query, the history-derived bias remains fixed and the control-derived clear-sky index is recomputed.

\begin{equation}\label{eq:4.3}\begin{adjustbox}{max width=.92\linewidth}$\displaystyle {\hat{y}}_{t+1:t+H}=F({x}_{t-L+1:t},{c}_{t-L+1:t},{c}_{t+1:t+H}).$\end{adjustbox}\end{equation}

where $F$ predicts active power on the original physical scale, giving the sequence ${\hat{y}}_{t+1:t+H}$ for the next $H$ time steps. Agreement with observations is measured by ${R}^{2}$:

\begin{equation}\label{eq:4.4}\begin{adjustbox}{max width=.92\linewidth}$\displaystyle R^{2}=1-\frac{\sum_{i\in\mathcal I}(y_i-\hat y_i)^{2}}{\sum_{i\in\mathcal I}(y_i-\bar y_{\mathcal I})^{2}}.$\end{adjustbox}\end{equation}

The set $\mathcal I$ comprises forecast--observation pairs for the specified station and horizon. For each pair, $y_i$ is observed active power and $\hat y_i$ is predicted active power; $\bar y_{\mathcal I}$ is the mean of the observations. Power forecasts support scheduling. Integrated energy and intervention responses are used for scenario assessment, which requires physical response fidelity alongside forecast accuracy. The model must combine inputs of different dimensions, limit error accumulation and preserve credible responses to weather changes.

\par\vspace{4pt}\noindent\begin{minipage}{\linewidth}
\refstepcounter{table}{\small\textbf{Table \thetable. Input Categories and Outputs of the Forecasting Tasks}}\label{tab:inputs}\par\vspace{4pt}
{\fontsize{8.5}{10.2}\selectfont\setlength{\tabcolsep}{3pt}
\begin{tabular}{>{\raggedright\arraybackslash}p{\dimexpr 0.15915\linewidth-2\tabcolsep\relax}>{\raggedright\arraybackslash}p{\dimexpr 0.37525\linewidth-2\tabcolsep\relax}>{\raggedright\arraybackslash}p{\dimexpr 0.46560\linewidth-2\tabcolsep\relax}}
\toprule
\textbf{Category} & \textbf{Wind (10 min)} & \textbf{PV (15 min)} \\
\midrule
History $x$ & Historical power, local meteorological observation and turbine availability; 3 channels, $L=24$ & Historical power, local meteorological observations and measured clear-sky index; 8 channels \\
Driver $c$ & Meteorological drivers; 5 channels per step & Meteorological forecasts and cyclic local-time features (10 history channels). The future-input sequence has 13 channels, including lagged power and derived irradiance features. \\
Output $\hat{y}$ & Active power ${\hat{y}}_{t+1:t+H}\in\mathbb{R}^{H}$ & Active power ${\hat{y}}_{t+1:t+H}\in\mathbb{R}^{H}$; auxiliary irradiance sequence \\
\bottomrule\end{tabular}}\end{minipage}\par\vspace{4pt}

\par\vspace{4pt}\noindent\begin{minipage}{\linewidth}
\refstepcounter{table}{\small\textbf{Table \thetable. Domain-specific implementations and training configurations.}}\label{tab:domain}\par\vspace{4pt}
{\fontsize{8.1}{9.7}\selectfont\setlength{\tabcolsep}{3pt}
\begin{tabular}{>{\raggedright\arraybackslash}p{0.185\linewidth}>{\raggedright\arraybackslash}p{0.385\linewidth}>{\raggedright\arraybackslash}p{0.385\linewidth}}
\toprule
\textbf{Component} & \textbf{PV implementation} & \textbf{Wind v4h implementation} \\
\midrule
Temporal input & 15-min steps; 18 history and 13 future-control channels & 10-min steps; 24-step history; 3 history and 5 drive channels \\
Encoder and state & Patch length 4; Transformer width 128, 2 layers, 4 heads; attention pooling; $d_z=64$ & One latent state partitioned as $[z_{\mathrm{phys}},z_{\mathrm{bias}}]$; $z_{\mathrm{bias}}$ encoded once from observation history \\
Transition and decoder & GRU transition; analytic irradiance--temperature--power chain with learned site correction and auxiliary irradiance output & GRU transition; multiplicative effective-wind gain followed by analytic power conversion and operating gates \\
Training objective & Power MSE $+$ JEPA representation loss ($\lambda_{\mathrm{repr}}=0.3$) $+$ sign regularization ($0.05$); night-time clamp is structural & Power MSE $+$ invariance ($20$) $+$ subspace decorrelation ($0.1$) $+$ monotonicity ($0.5$) \\
Training and selection & Four-site repeatability experiment: horizon-specific pretrained initialization; 40 epochs; learning rate $10^{-3}$; independent seed-specific training & Random initialization; at most 60 epochs; EMA (exponential moving average) stopping ($\alpha=0.6$), 5-epoch burn-in and patience 15; lowest post-burn-in raw validation loss selected \\

\bottomrule
\end{tabular}}
\end{minipage}\par\vspace{4pt}

\tabref{tab:domain} summarizes the domain implementations. In the four-site PV repeatability experiment, the three runs for each horizon start from the same pretrained checkpoint and use different training seeds. The 4 h task uses the 4 h pretrained checkpoint; the 12 h and 1 d tasks use the 1 d checkpoint. The runs use a stride of eight and share the same pretrained initialization for each horizon, with different seeds used for subsequent training. The matched v4e and v4h wind models are trained from random initialization for at most 60 epochs. These initialization settings apply to the stated experiments and are not assumed for every PV ablation. For the six-site response summaries, $s$, $h$ and $r$ index site, horizon and training seed. Let $\alpha=1.2$ denote the prescribed global horizontal irradiance (GHI) scaling, and let $\rho_{s,h,r}(\alpha)$ denote the net per-unit response for one run. The response over the valid time points is
\begin{equation}\label{eq:rho}
\rho_{s,h,r}(\alpha)=\frac{1}{|\mathcal{T}_{s,h,r}|}\sum_{t\in\mathcal{T}_{s,h,r}}
\frac{\hat y^{\mathrm{cf}}_{s,h,r,t}-\hat y^{\mathrm{base}}_{s,h,r,t}}
{\max\!\left(\hat y^{\mathrm{base}}_{s,h,r,t},\epsilon\right)}\frac{1}{\alpha-1},\qquad \alpha=1.2,\quad \epsilon=10^{-6},
\end{equation}
where $\mathcal{T}_{s,h,r}=\{t:0.05P_{\mathrm{cap}}<\hat y^{\mathrm{base}}_{s,h,r,t}<0.90P_{\mathrm{cap}}\}$ and a window is scored only when $|\mathcal{T}_{s,h,r}|\geq 20$. Thus, $\rho$ is the mean relative net response per unit irradiance intervention, rather than the raw counterfactual-to-baseline power ratio. The ideal unit-proportional response is $\rho=1$; the site- and horizon-specific analytic expectations range from 0.922 to 0.981 because the slowly varying loss factor is retained in the analytic response chain. The summaries in Supplementary \tabref{tab:s3} use

\begin{equation}\label{eq:rho-aggregate}\tilde\rho_{s,h}=\operatorname{median}_{r\in\{0,1,2\}}\rho_{s,h,r},\qquad \Delta\rho_{s,h}=\tilde\rho_{s,h}-\rho^{\mathrm{analytic}}_{s,h}.\end{equation}

\eqnref{eq:rho-aggregate} specifies aggregation across seeds and comparison with the analytic reference. The mean is taken over valid time points before the three seed-level scores are summarized by their median; a pointwise median would saturate at 1.0 for two sites and lose resolution. Excluding clipping-saturated points can make $\rho$ slightly exceed the corresponding analytic expectation because those points would otherwise pull the response ratio down. Within each reported comparison, $\Delta\rho=0$ denotes agreement with its analytic reference, not unchanged power. \figfullref{fig:decoders} and Supplementary \tabref{tab:s2} average 4 h and 1 d scores from one seed; \figfullref{fig:interventions}A and \tabref{tab:s6} use seed 0 at 1 d. These summaries are distinct from the three-seed medians in \tabref{tab:s3}.

EWM was evaluated for forecasting, training stability, decoder contributions and intervention response. PV covered operational forecasts at four sites and response tests at six sites at 4 h, 12 h and 1 d. Wind comprised matched Kelmarsh 4 h controls. Independent training runs indexed by seeds 0, 1 and 2 share the data split and follow the initialization protocol specified above for each experiment. $R^2$ was computed on the original power scale for each stated site and horizon; replicated results are reported as the mean $\pm$ sample standard deviation (SD) across three independent training runs using the same data split. The experimental scope is summarized in Supplementary Section S0; the retained supplementary tables are cited at the corresponding points in the main text.

The principal wind model uses hybrid checkpoint selection, in which an exponential moving average (EMA) of validation loss controls early stopping while the lowest raw post-burn-in validation loss determines the final checkpoint. EMA smoothing is set to 0.6, with a five-epoch burn-in and an early-stopping patience of 15 epochs. A single checkpoint is selected, and the test set is used only for final evaluation. Comparisons with EMA-based selection use identical seed-specific training trajectories to isolate the effect of checkpoint selection on forecasting performance and between-seed variability. Wind-injection elasticity is $\varepsilon_v=d\log v_{\mathrm{eff}}/d\log s$, where $s$ scales the NWP wind-speed input and $v_{\mathrm{eff}}$ is the model-internal effective wind speed. It is estimated over $s\in\{0.4,0.6,0.8,1.0,1.2,1.5\}$ on 64 windows distributed across the test split under deterministic rollout; the target is $1.000\pm0.02$. The pre-window timing test separately requires a perturbation beginning at step $a$ to leave predictions at steps $j<a$ unchanged within the archived numerical tolerance.

\subsection*{4.2 Forecasting performance and training stability}

\subsubsection*{4.2.1 PV forecasting performance and stability}

\begin{figure}[!t]\centering
\includegraphics[width=\linewidth]{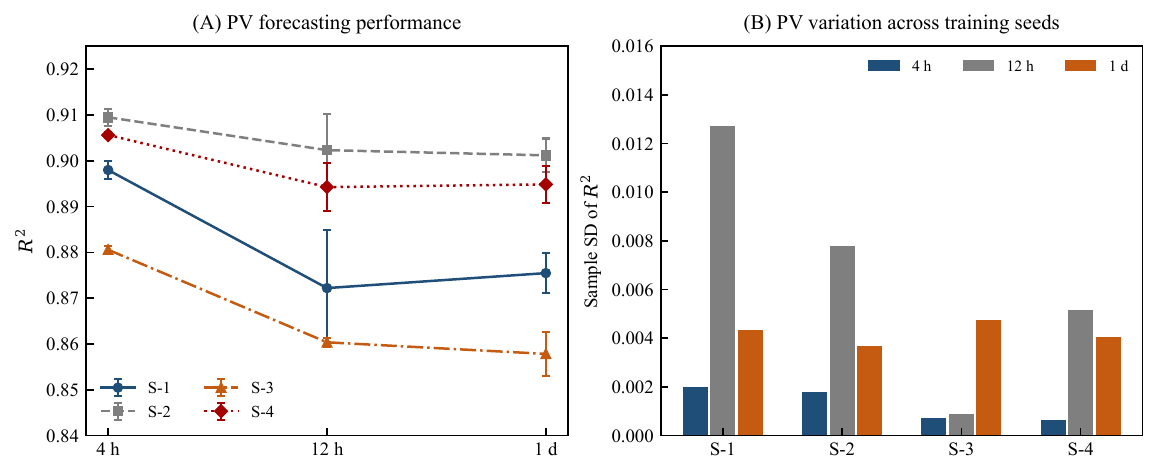}
\caption{PV forecasting performance and repeatability. A, Mean R\ensuremath{^{2}} \ensuremath{\pm} sample SD at four sites and three horizons; each point summarizes three seeds. B, Sample SD of R\ensuremath{^{2}} for the same site–horizon combinations; lower bars indicate closer agreement between repeated training runs.}\label{fig:pv-forecast}
\end{figure}

Across the 12 site--horizon combinations, mean R\ensuremath{^{2}} ranged from 0.8578 to 0.9095 (\figref{fig:pv-forecast}A). The pooled means were 0.8984, 0.8823 and 0.8823 at 4 h, 12 h and 1 d, respectively. Complete values are reported in Supplementary \tabref{tab:s1}.

At S-2, R\ensuremath{^{2}} reached 0.9095 $\pm$ 0.0018 at 4 h and remained 0.9012 $\pm$ 0.0037 at 1 d, showing sustained fit at the longer horizon. \figfullref{fig:pv-forecast}A also reports all other sites, including S-4, whose horizon-specific means ranged from 0.8942 to 0.9056. The three-seed means and error bars summarize forecast performance and its variability across repeated training runs.

As shown in \figref{fig:pv-forecast}B, repeated PV training produced small dispersion in most tasks: sample SD of R\ensuremath{^{2}} was below 0.006 in 10 of the 12 site--horizon combinations, and the largest 4 h SD was approximately 0.0020. After averaging each seed across all 12 site--horizon cells, the three seed-level global means spanned only 0.00049. Local dispersion was greater for S-1 at 12 h, where SD was approximately 0.0127. Thus, stable aggregate performance did not eliminate site- and horizon-dependent variation.

Performance decreased from 4 h to 12 h at all four sites, whereas the pooled mean changed little between 12 h and 1 d. The effect of extending the horizon therefore depended on the site and horizon rather than following a uniform linear trend. Repeated training separately quantified variation associated with the training seed.

\subsubsection*{4.2.2 Wind forecasting performance and stability}

\begin{figure}[!t]\centering
\includegraphics[width=\linewidth]{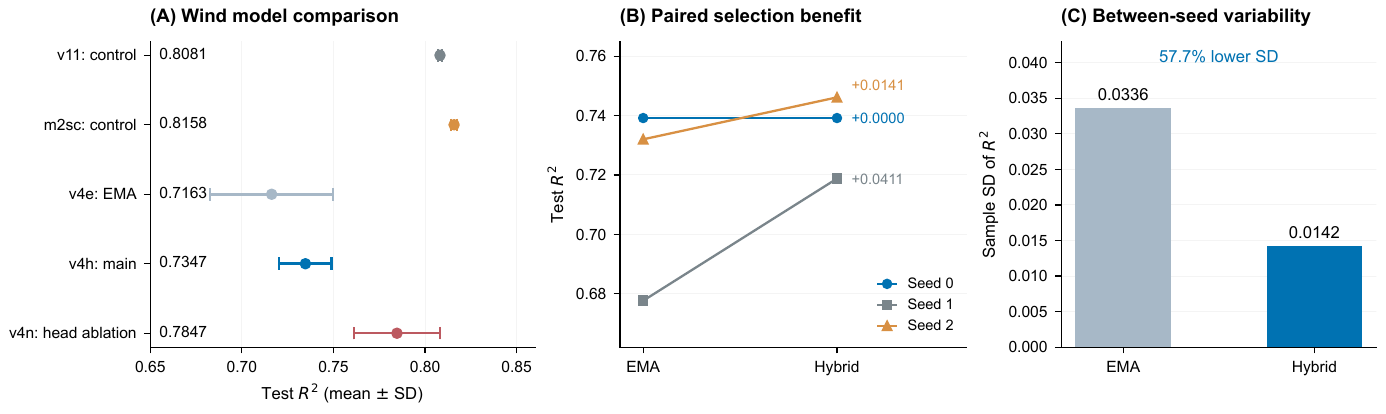}
\caption{Wind model comparison and checkpoint-selection benefits at Kelmarsh, 4 h. A, Mean R\ensuremath{^{2}} and sample SD across three seeds for each model: v11 and m2sc are structural controls, v4e is the EMA-selection control, v4h is the principal model and v4n is the power-head ablation. B, EMA and hybrid selection on identical seed-specific training trajectories; labels give changes in R\ensuremath{^{2}}. C, Between-seed sample SD of R\ensuremath{^{2}}, reduced by 57.7\% with hybrid selection. This SD measures between-run variability in forecast accuracy; convergence speed is a separate training characteristic.}\label{fig:wind}
\end{figure}

The principal offline EWM configuration is v4h, which uses unified physical decoding; seed-level results are reported in Supplementary \tabref{tab:s5}. The v11 and m2sc models are structural controls, v4e uses the same unified structure with EMA checkpoint selection, and v4n replaces only the power head. In \figref{fig:wind}A, v4n increases mean R\ensuremath{^{2}} from 0.7347 $\pm$ 0.0142 for v4h to 0.7847 $\pm$ 0.0236. This ablation identifies the power mapping as a principal source of the remaining fit gap rather than replacing v4h as the principal configuration.

\par\vspace{4pt}\noindent\begin{minipage}{\linewidth}
\refstepcounter{table}{\small\textbf{Table \thetable. Forecasting and physical-response results for matched wind controls at Kelmarsh, 4 h.}}\label{tab:wind-controls}\par\vspace{4pt}
{\fontsize{8.5}{10.2}\selectfont\setlength{\tabcolsep}{3pt}
\begin{tabular}{>{\raggedright\arraybackslash}p{\dimexpr 0.22741\linewidth-2\tabcolsep\relax}>{\raggedright\arraybackslash}p{\dimexpr 0.19332\linewidth-2\tabcolsep\relax}>{\raggedright\arraybackslash}p{\dimexpr 0.19332\linewidth-2\tabcolsep\relax}>{\raggedright\arraybackslash}p{\dimexpr 0.28988\linewidth-2\tabcolsep\relax}>{\raggedright\arraybackslash}p{\dimexpr 0.09606\linewidth-2\tabcolsep\relax}}
\toprule
\textbf{Model and role} & \textbf{R\ensuremath{^{2}} mean \ensuremath{\pm} SD $\uparrow$} & \textbf{R\ensuremath{^{2}} change and reference} & \textbf{$\varepsilon_v$ mean \ensuremath{\pm} SD; target $=1$} & \textbf{pre-window timing} \\
\midrule
v11 Bidirectional control & 0.8081 \ensuremath{\pm} 0.0013 & — & N/E & FAIL \\
m2sc Additive control & 0.8158 \ensuremath{\pm} 0.0017 & — & 0.2367 \ensuremath{\pm} 0.0238 & PASS \\
v4e EMA-selection control & 0.7163 \ensuremath{\pm} 0.0336 & — & 0.9991 \ensuremath{\pm} 0.0004 & PASS \\
v4h Principal offline EWM & 0.7347 \ensuremath{\pm} 0.0142 & +0.0184 vs v4e & 0.9990 \ensuremath{\pm} 0.0006 & PASS \\
v4n Power-head ablation & 0.7847 \ensuremath{\pm} 0.0236 & +0.0500 vs v4h & 0.9999 \ensuremath{\pm} 0.0003 & PASS \\
\bottomrule\end{tabular}}\end{minipage}\par\vspace{4pt}

Note: Means and sample SD are calculated from three training runs; v4n values retain the rounding in the source. Higher R\ensuremath{^{2}} indicates a closer fit to observed power. An $\varepsilon_v$ near one indicates proportional transmission of an imposed wind-speed change to effective wind speed, with little attenuation or amplification. For a perturbation starting at step $a$, the timing test checks whether earlier outputs remain unchanged: PASS denotes an intact temporal boundary and FAIL a violation. In v11, bidirectional temporal coupling allowed future-driver changes to affect pre-intervention outputs (\secreftext{sec:wind-intervention}{Section~4.4.2}). N/E means not evaluated. All controls were evaluated in this study.

Predictive fit and intervention-response fidelity are reported separately in \tabref{tab:wind-controls}. Under the matched Kelmarsh evaluation, v4h achieved a wind-injection elasticity of 0.9990 $\pm$ 0.0006, compared with 0.2367 $\pm$ 0.0238 for m2sc. Thus, v4h transmitted the imposed changes in issued wind speed almost proportionally to effective wind speed in the evaluated perturbation and scenario rollouts. This response-fidelity advantage was accompanied by lower forecasting R\ensuremath{^{2}} than m2sc, which reached 0.8158 $\pm$ 0.0017. Relative to the v4h--m2sc gap of $0.8158-0.7347=0.0811$, the v4n gain of 0.0500 recovered 61.6\% while retaining near-unit elasticity, localizing the main opportunity for improvement to the power decoder. The v4h results were obtained offline, and the configuration had not been deployed at the time of evaluation; the operational wind model belonged to the m2sc/v11 lineage.

As shown in \figref{fig:wind}B, hybrid selection improved test performance for two seeds and retained the existing result for the third, increasing mean R\ensuremath{^{2}} from 0.7163 to 0.7347. The lowest-performing seed under the original selection improved from 0.6776 to 0.7188. Because the architecture and each seed's training trajectory were unchanged, this gain can be attributed to checkpoint selection.

Together, \figfullref{fig:wind}B and \figfullref{fig:wind}C show that validation-based selection of a single checkpoint increased mean R\ensuremath{^{2}} by 0.0184 and reduced between-seed SD by 57.7\%. The selection protocol in \secreftext{sec:case-study}{Section~4.1} uses validation loss for both early stopping and final checkpoint selection.

\subsection*{4.3 Decoder contributions and physical interpretation}\phantomsection\label{sec:decoder-contributions}

\subsubsection*{4.3.1 PV physical decoder}

Here, ``physical decoder'' denotes the framework module, ``analytic physical decoder'' denotes the PV case-specific implementation, and ``MLP decoder'' denotes the learned comparator.

The principal PV model uses an analytic physical decoder linking irradiance, cell temperature, efficiency derating and power, together with a learned site correction \ensuremath{\eta}. This factor compensates for conversion discrepancies not represented by the analytic chain and is intended to represent the combined effect of slowly varying losses, including soiling, mismatch and degradation. The MLP ablation is a multilayer perceptron replacing the analytic conversion with a learned mapping. The comparison therefore isolates the output module within the same forecasting framework. The net per-unit response \ensuremath{\rho} expresses the response to a prescribed irradiance change and is compared with the corresponding site- and horizon-specific analytic reference. The case-specific PV mapping used in the decoder comparison is summarized by

\begin{equation}\label{eq:4.6}\begin{adjustbox}{max width=.92\linewidth}$\displaystyle {\hat y}^{\mathrm{PV}}_{t+j}=\eta({z}_{t+j})\,P_{\mathrm{analytic}}({G}_{t+j},{T}_{\mathrm{cell},t+j}),\qquad {G}_{t+j}\rightarrow {T}_{\mathrm{cell},t+j}\rightarrow {r}_{T,t+j}\rightarrow {P}_{t+j}.$\end{adjustbox}\end{equation}

Here $G$ is irradiance, $T_{\mathrm{cell}}$ is cell temperature, $r_T$ is the temperature-derating factor and $\eta$ is a learned site correction. The correction is intended to encode unresolved slowly varying effects, but the architecture does not by itself guarantee invariance to a queried irradiance change. This equation is used only for the PV case-study decoder and response diagnostics.

\begin{figure}[!t]\centering
\includegraphics[width=\linewidth]{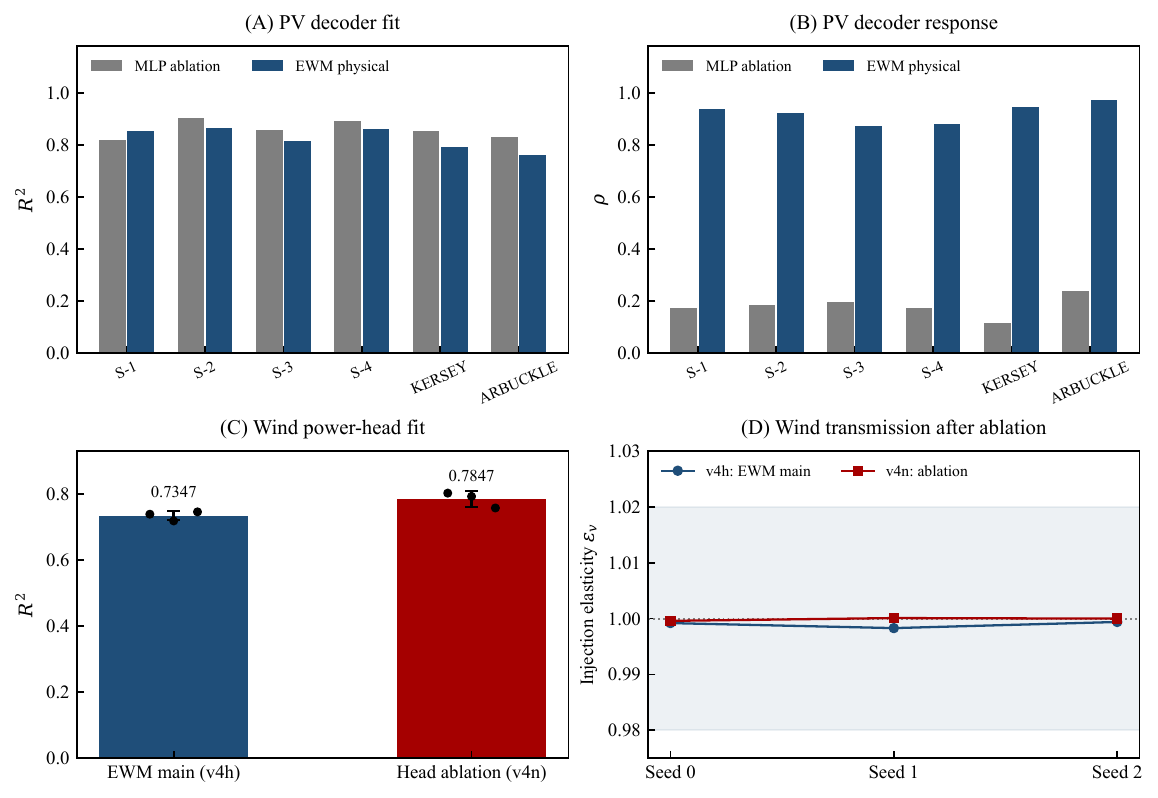}
\caption{Decoder contributions to fit and response. A,B, EWM physical PV decoder in blue and MLP ablation in grey; values average 4 h and 1 d results from one seed, with no seed error bars. Higher R\ensuremath{^{2}} indicates closer agreement with observations, whereas \ensuremath{\rho} is assessed by proximity to its analytic expectation. C, Principal wind model v4h in blue and power-head ablation v4n in red; bars show R\ensuremath{^{2}} mean \ensuremath{\pm} sample SD and dots individual runs. D, $\varepsilon_v$ for both versions, with the 1.00 \ensuremath{\pm} 0.02 target band; values nearer one indicate more faithful wind-speed transmission. PV response scores use the net per-unit definition in \eqnref{eq:rho} and are interpreted relative to the corresponding analytic expectation.}\label{fig:decoders}
\end{figure}

\figfullref{fig:decoders}A evaluates forecast fit with analytic physical decoding, with higher R\ensuremath{^{2}} indicating closer agreement with observations. With the encoder and training recipe fixed, the six-site, single-seed ablation gave a mean R\ensuremath{^{2}} of 0.825 for the EWM physical decoder. The explicit conversion relationship therefore retained the capacity to fit observed power variations. At S-1, R\ensuremath{^{2}} increased from 0.818 with the MLP to 0.853 with physical decoding. Across all six sites, the MLP mean was higher at 0.859, consistent with the flexibility of a learned output mapping. \figfullref{fig:decoders}B compares the decoders' irradiance responses to assess their use in environmental scenario analysis.

The net per-unit response \ensuremath{\rho} measures how an irradiance perturbation propagates to predicted power. Its agreement with the physical relationship is judged against the analytic reference for that site and forecast horizon. In \figfullref{fig:decoders}B, the analytic physical decoder increased the reported \ensuremath{\rho} at all six sites, raising the mean from 0.180 for the MLP to 0.922, a difference of 0.742. Site-level scores ranged from 0.871 to 0.974 for the physical decoder and from 0.117 to 0.238 for the MLP. This consistent cross-site pattern is consistent with reduced attenuation of irradiance perturbations through the analytic conversion chain and supports its use in EWM irradiance-scenario analysis. The score remains a diagnostic for the prescribed intervention and is interpreted relative to the analytic reference.

\figfullref{fig:decoders}A and \figfullref{fig:decoders}B assess complementary properties. R\ensuremath{^{2}} measures fit under the observed weather conditions, whereas $\rho$ measures whether the predicted output responds appropriately after the driver is changed. The physical decoder increased mean $\rho$ by 0.742 and reduced response attenuation at all six sites while retaining competitive forecast fit. Because a short weather query should not immediately change slowly varying losses, dependence of $\eta$ on queried irradiance can still offset part of the analytic response; \secreftext{sec:pv-intervention}{Section~4.4.1} tests this mechanism.

Together, the two panels motivate separate evaluation of forecast fit and driver response. The analytic decoder provides an explicit physical conversion relationship and a reference for response assessment, extending model evaluation from forecasting to environmental scenario queries.

\subsubsection*{4.3.2 Wind gain layer and power decoder}\phantomsection\label{sec:wind-decoder}

The wind decoder separated effective-wind-speed correction from power conversion. A gain layer multiplied external wind speed by corrections derived from static bias and dynamic state. The power layer then used air density, swept area, conversion efficiency and the cubic wind-speed relationship, together with rated-capacity and cut-in/cut-out constraints. Effective wind speed is an internal corrected drive, not an independently measured quantity. Its role is to connect external wind conditions with turbine conversion characteristics and allow response errors to be localized by module. The case-specific latent partition and effective-speed mapping are

\begin{equation}\label{eq:4.7}\begin{adjustbox}{max width=.92\linewidth}$\displaystyle {z}_{t+j}=[{z}_{\mathrm{phys},t+j},{z}_{\mathrm{bias}}],\qquad {v}_{\mathrm{eff},t+j}={ws}_{t+j}\exp\!\left({g}_{\mathrm{static}}({z}_{\mathrm{bias}})+{g}_{\mathrm{mem}}({z}_{\mathrm{phys},t+j})\right).$\end{adjustbox}\end{equation}

The aerodynamic conversion is

\begin{equation}\label{eq:4.8}\begin{adjustbox}{max width=.92\linewidth}$\displaystyle {P}_{\mathrm{aero},t+j}=\frac{1}{2}\rho_{\mathrm{air}} A{C}_{p}{v}_{\mathrm{eff},t+j}^{3},\qquad \hat y^{\mathrm{wind}}_{t+j}=\mathcal G_{\mathrm{op}}\!\left(\kappa_{\mathrm{site}}P_{\mathrm{aero},t+j}\right).$\end{adjustbox}\end{equation}

Here $ws$ is issued NWP wind speed; $z_{\mathrm{bias}}$ is a history-only static code; $z_{\mathrm{phys}}$ is the rolled physical subspace; and $g_{\mathrm{static}}$ and $g_{\mathrm{mem}}$ are their gain mappings. The exponential enforces a positive multiplicative gain. In \eqnref{eq:4.8}, $\rho_{\mathrm{air}}$ is air density, $A$ is rotor-swept area, $C_p$ is the power coefficient, $\kappa_{\mathrm{site}}$ is the learned site factor and $\mathcal G_{\mathrm{op}}$ applies rated-capacity and soft cut-in/cut-out gates. Because the queried wind level enters $v_{\mathrm{eff}}$ through the explicit multiplicative factor $ws_{t+j}$, while the static code is fixed and the dynamic gain excludes direct wind-level readout, the architecture targets $\varepsilon_v\simeq1$.

The wind transition acts on the dynamic subspace as $z_{\mathrm{phys},t+j}=\operatorname{GRUCell}(\operatorname{ctrlproj}_{\mathrm{wind}}(c_{t+j}),z_{\mathrm{phys},t+j-1})$, while $z_{\mathrm{bias},t+j}=z_{\mathrm{bias},t}$. Here $\operatorname{ctrlproj}_{\mathrm{wind}}$ is the domain-specific drive projection: it excludes the absolute wind-level channel and admits the prescribed deviation-structured features. The static code is encoded once from observation history and is shared by the baseline and query throughout the horizon. Thus $T$ in \eqnref{eq:3.3} combines a recurrent update with identity propagation of the static component. A new observation window can change that component at the next issue time; a future-drive query does not re-encode it. This is a partition within one model, not a switch between forecasting and query models.

\figfullref{fig:decoders}C isolates the contribution of the wind power decoder by replacing only the analytic power head of v4h to obtain v4n. R\ensuremath{^{2}} increased in all three runs, from 0.7347 $\pm$ 0.0142 to 0.7847 $\pm$ 0.0236 on average, while the upstream driver-transmission path was unchanged. The result is consistent with a substantial contribution from the effective-wind-to-power mapping to the remaining fit gap. In the tested range, v4n retained cut-out gating and reached a plateau near 0.993 of rated output. The analytic chain reached approximately 0.850 because of its learned site factor. This difference points to the effective-wind-to-power mapping as a target for decoder refinement under the evaluated wind conditions.

Replacing the power head left wind-injection elasticity $\varepsilon_v$ close to one (\figref{fig:decoders}D): the three learned-head runs gave 0.9996, 1.0001 and 1.0000. A proportional change in external wind speed thus produced nearly the same proportional change in effective wind speed. The downstream power mapping remained nonlinear, with cubic growth below rated operation, saturation and cut-out decline.

The wind ablation identifies distinct roles for the gain path and the power head. Gain-path constraints preserve injection amplitude; the power head determines forecast fit and curve shape. Static bias comes from unperturbed observation history, while the dynamic gain restricts direct readout of wind level to reduce arbitrary cancellation of the imposed change. Refinement can therefore focus on the power mapping while preserving the near-unit upstream transmission measured under the evaluated perturbations.

\subsection*{4.4 Intervention response and inversion capability}\phantomsection\label{sec:intervention}

\subsubsection*{4.4.1 PV intervention response}\phantomsection\label{sec:pv-intervention}

\begin{figure}[!t]\centering
\includegraphics[width=\linewidth]{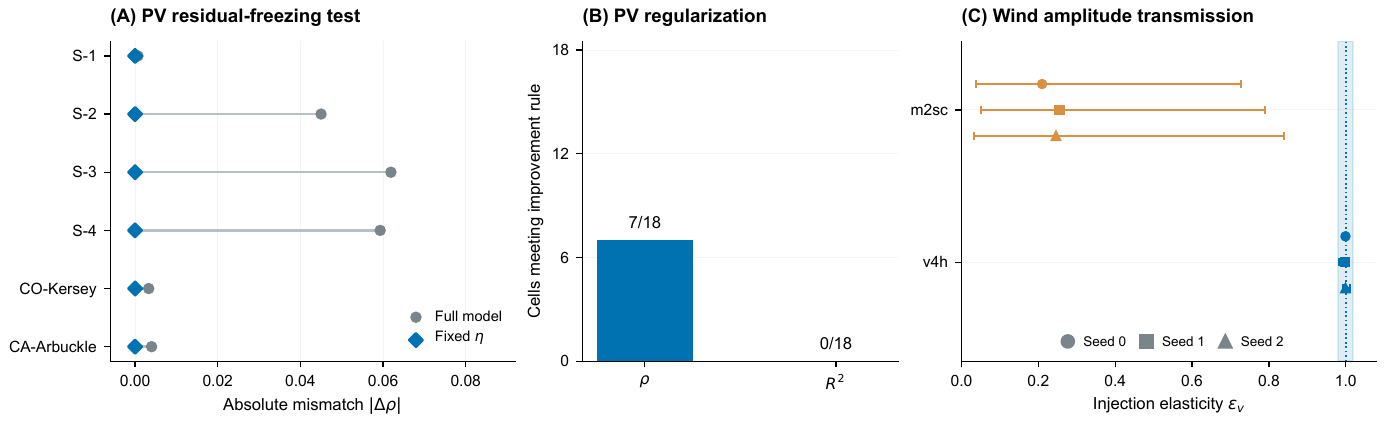}
\caption{Response consistency under prescribed weather interventions. A, Absolute PV response mismatch |\ensuremath{\Delta}\ensuremath{\rho}| for the full model and fixed \ensuremath{\eta} under GHI\ensuremath{\times}1.2, seed 0, 1 d. Paired points correspond to the same site; lower is closer to the analytic expectation. Zero mismatch for fixed \ensuremath{\eta} is limited to the four-decimal precision of the recorded data. B, Task cells meeting the source improvement rule out of 18; the rule is a screening criterion rather than a significance test. C, Seed-mean $\varepsilon_v$ and extrema over 64 windows for wind scales $\{0.4,0.6,0.8,1.0,1.2,1.5\}$; lines are observed ranges rather than confidence intervals. Shading indicates the $1.00\pm0.02$ target band. PV panels report the net per-unit diagnostic defined in \eqnref{eq:rho}, interpreted relative to the analytic expectation for the prescribed intervention.}\label{fig:interventions}
\end{figure}

\figfullref{fig:interventions}A examines PV response error in the six-site, seed-0, 1 d diagnostic. Global horizontal irradiance (GHI) is increased by 20\% (\ensuremath{\times}1.2) in both the full model and a control with fixed \ensuremath{\eta}. In the control, each forecast step retains its time-matched, pre-intervention site correction. The signed mismatch \ensuremath{\Delta}\ensuremath{\rho} is the measured response minus the analytic expectation; negative values indicate attenuation and zero indicates agreement. The panel plots |\ensuremath{\Delta}\ensuremath{\rho}|, so smaller values denote closer agreement. With \ensuremath{\eta} fixed, all six sites match their analytic expectations to the four decimal places reported in Supplementary \tabref{tab:s6}.

A short-term irradiance increase is expected to propagate first through the specified energy-conversion relationship. Freezing $\eta$ separates that weather change from the slowly varying site correction by retaining the time-matched baseline correction at every forecast step. This control tests a structural explanation for the response deficit and identifies residual input dependence as a target for refinement. Frozen-state diagnostics are reported separately from full-model results under operational conditions.

\figfullref{fig:interventions}B evaluates regularization in the full model with \ensuremath{\eta} unfrozen. The six sites and three horizons give 18 task cells. A cell is counted as improved when all three runs improve in the same direction and the change exceeds the archived dispersion threshold. Regularization meets this response-improvement rule in seven cells and the R\ensuremath{^{2}}-improvement threshold in none; the rule is a screening criterion rather than a significance test. The complete task cells are given in Supplementary \tabref{tab:s3}. At S-4, the unregularized $\Delta\rho$ range of approximately $[-0.112,-0.090]$ contracts to the regularized range $[-0.064,-0.059]$. The observed gain is reduced response cancellation and improved physical consistency.

The archived screening thresholds were $2\hat\sigma=0.0078$ for $R^2$ and $0.038$ for $\rho$. The dispersion estimate was $\hat\sigma=\operatorname{median}(\mathrm{range})/d_2(3)$, where each range spans the three seed results in a task cell and $d_2(3)$ is the range-to-standard-deviation conversion factor for three observations. These thresholds summarize seed dispersion and are not inferential significance levels.

\subsubsection*{4.4.2 Wind intervention response}\phantomsection\label{sec:wind-intervention}

Wind-response fidelity remained stable across the prescribed perturbation range (\figref{fig:interventions}C). Under deterministic rollout, scaling wind speed over $\{0.4,0.6,0.8,1.0,1.2,1.5\}$ on 64 windows distributed across the full test split produced seed-mean $\varepsilon_v$ of approximately 0.998--0.999 for v4h, with individual-window values spanning approximately 0.984--1.010. In contrast, m2sc produced seed means of only 0.210--0.255 and greater variation across windows. The unified gain path therefore transmitted imposed wind-speed changes more completely and consistently within the evaluated range.

Temporal consistency and amplitude transmission capture distinct properties of a scenario response. The pre-window timing test requires a future wind-speed perturbation to leave predictions before its onset unchanged. The bidirectional temporal coupling in v11 allowed the modified future-driver segment to affect pre-intervention outputs, producing the FAIL result in \tabref{tab:wind-controls}; this result concerns suitability for driver-substitution queries rather than ordinary forecasting accuracy. The m2sc, v4h and v4n configurations passed the timing test. Their amplitude responses differed: m2sc attenuated the imposed wind change, whereas v4h retained near-unit elasticity, which v4n preserved after power-decoder replacement. A scenario response therefore requires assessment of both timing and amplitude. No elasticity record is available for v11, and these findings apply only to the tested perturbations and windows.

For the prescribed weather changes, wind transmission varied little across the evaluated windows and seeds, and the PV analytic chain and regularization reduced response deficits. These findings support the consistency of model responses used for weather-scenario queries within the tested driver ranges. Establishing causal effects of weather changes at real plants requires evidence beyond these model-based experiments.

\subsubsection*{4.4.3 Site coverage and target inversion}\phantomsection\label{sec:inversion}

Response deficits vary by site and horizon in the six-site PV audit (Supplementary \tabref{tab:s3}). S-1 and P-CA-ARBUCKLE stay close to the analytic expectation across horizons, while S-3 and S-4 have larger negative $\Delta\rho$. At P-CO-KERSEY, the signed $\Delta\rho$ changes from $-0.0541$ at 4 h to $-0.0039$ at 1 d, indicating a smaller response deficit. The full matrix shows this local variation and specifies the sites and horizons over which the response claim is supported.

The target-inversion query seeks an irradiance-drive scale $s^*$ that makes $E(s^*)/E(1)$ match the requested energy ratio $r^*$. Here $E(s)$ is the energy obtained by applying the drive scale $s$ and integrating the resulting forecast over the same window $W$ as in \eqnref{eq:3.8}; $E(1)$ is the valid, non-zero baseline energy. A target of $r^*=1.10$, for example, requests a 10\% increase in predicted energy. The inverse query determines the corresponding $s^*$ while keeping the initial state, model weights, other drivers and integration window fixed. Back-substitution checks the result against the same forward model. The archived queries cover 12 h and 1 d targets in $[0.9,1.1]$; Supplementary Section S4 defines the query and its reporting scope. Quantitative inversion accuracy remains outside the present evidence because the error normalization and solver settings are not specified.

\subsection*{4.5 Applicability and performance differences across energy domains}

\tabref{tab:cross-domain} compares the PV and wind evidence across four dimensions: power-forecast fit, consistency across training runs, decoder behavior and intervention response. The latter two test how weather changes propagate through the model to its output. PV operational results span multiple sites and horizons; matched wind experiments examine checkpoint selection and power-head design. Both implementations follow the same computational framework with domain-specific parameters, and each is evaluated against controls from its own domain.

\par\vspace{4pt}\noindent\begin{minipage}{\linewidth}
\refstepcounter{table}{\small\textbf{Table \thetable. Corresponding PV and wind findings and algorithmic contributions.}}\label{tab:cross-domain}\par\vspace{4pt}
{\fontsize{8.5}{10.2}\selectfont\setlength{\tabcolsep}{3pt}
\begin{tabular}{>{\raggedright\arraybackslash}p{\dimexpr 0.25000\linewidth-2\tabcolsep\relax}>{\raggedright\arraybackslash}p{\dimexpr 0.25000\linewidth-2\tabcolsep\relax}>{\raggedright\arraybackslash}p{\dimexpr 0.25000\linewidth-2\tabcolsep\relax}>{\raggedright\arraybackslash}p{\dimexpr 0.25000\linewidth-2\tabcolsep\relax}}
\toprule
\textbf{Dimension} & \textbf{PV evidence} & \textbf{Wind evidence} & \textbf{Supported benefit} \\
\midrule
Fit and repeatability & Four-site R\ensuremath{^{2}}: 0.8578–0.9095; SD<0.006 in 10/12 cells & v4h vs EMA: R\ensuremath{^{2}} +0.0184; SD \ensuremath{-}57.7\% & Consistent performance over evaluated horizons; more reproducible validation selection \\
Decoder response & Physical chain vs MLP: mean \ensuremath{\rho} +0.742; all six sites improve & v4h vs m2sc: $\varepsilon_v$ 0.237 to 0.999 & Less drive attenuation; preserved weather-change amplitude \\
Error localization & Fixed \ensuremath{\eta} matches analytic response at six sites & v4n vs v4h: mean R\ensuremath{^{2}} +0.0500, $\varepsilon_v$ remains near one & Localizes residual coupling and power-mapping constraints \\
Query scope & Target-to-drive inversion formulated for 12 h and 1 d; numerical accuracy not established & v4h preserves the timing boundary and near-unit drive transmission at Kelmarsh & Defines inverse queries for PV and supports evaluated forward queries for wind \\
\bottomrule\end{tabular}}\end{minipage}\par\vspace{4pt}

Note: Benefits use within-domain controls; \ensuremath{\rho} and $\varepsilon_v$ quantify different response stages and are interpreted against their respective physical references. The v4n result is an ablation, whereas v4h remains the principal model; fixed \ensuremath{\eta} is a diagnostic condition. PV response-based benefits use the net per-unit definition in \eqnref{eq:rho} and are limited to the prescribed intervention.

For forecasting, aggregate R\ensuremath{^{2}} in the present PV operational tasks was higher than that of the unified wind model at Kelmarsh, 4 h. However, datasets, horizons and input conditions differed, so the numerical gap alone cannot rank cross-domain model quality. PV power depends on irradiance and temperature. Wind power also involves nonlinear speed-to-power conversion across distinct operating regimes. These differences may affect forecasting difficulty and require analysis of the drive data. In the present wind experiments, the identified gains came from power-head replacement and checkpoint selection.

The response metrics examine different stages of the two models. PV \ensuremath{\rho} measures the net output response, and \ensuremath{\Delta}\ensuremath{\rho} its deviation from the analytic expectation. Wind $\varepsilon_v$ measures amplitude transmission from external to effective wind speed, with pre-window timing testing the temporal boundary. Both domains benefit from an explicit path for prescribed driver changes and checks on their transmission. The response improvements centered on the PV output conversion chain and the wind upstream injection path, supporting further refinement tailored to each module.

\subsection*{4.6 Limitations}

The evidence supports forecast and response performance at the evaluated sites and under the prescribed interventions. The matched evaluation of the unified wind model focused on Kelmarsh at a 4 h horizon; the PV decoder ablation used a single training seed. Future evaluation should address regional transfer, unseen-site performance, out-of-distribution weather, additional training repetitions, abrupt wind events, rapid cloud shading and input noise. The present queries therefore operate on explicit external drivers, while the physical interpretation of latent coordinates remains a target for further study.

The main model-level issues are local cancellation of PV weather responses by the residual and the trade-off between wind-power fit and response-preserving conversion constraints. Freezing and power-head ablations identify specific targets: constrain residual input dependence and refine the power mapping, then validate within adequately covered drive ranges. The v4h wind configuration was evaluated offline and had not been deployed at the time of evaluation. Reproducibility would benefit from fuller reporting of data splits, metric units and environment settings. Matched controls, repeated runs and structural ablations provide empirical tests of the constraints imposed by the architecture. The PV response diagnostics use the net per-unit definition in \eqnref{eq:rho}; numerical inversion accuracy requires a specified solver and error normalization.

\FloatBarrier

\papersection{Conclusion}\label{sec:conclusion}

An Earth-centred world model (EWM) is developed to unify multi-source information representation, recursive latent-state evolution, and physics-constrained decoding within a single dynamical framework. EWM uses explicit, replaceable environmental drivers for both multi-step forecasting and driver-substitution analysis. The evaluated model interface supports direct assessment of response timing and amplitude under prescribed interventions, providing a common framework for predictive modelling and environmental-response analysis of Earth-surface systems.

EWM is evaluated for photovoltaic and wind-power forecasting in terms of forecasting performance, physical decoding, and response to environmental perturbations. Across multiple forecast horizons at four photovoltaic sites, EWM achieved mean coefficients of determination ($R^2$) ranging from 0.8578 to 0.9095, with limited between-seed variation under the evaluated settings. Relative to multilayer-perceptron decoding, the analytic physical decoder increased the mean net per-unit response $\rho$ from 0.180 to 0.922, consistent with reduced attenuation of the prescribed irradiance change during output mapping. Holding the site correction factor fixed brought the responses at all six PV sites close to their analytic references. The wind model maintained the pre-intervention temporal boundary and achieved a wind-speed injection elasticity of 0.9990 $\pm$ 0.0006. These results indicate that predictive fit and physical-response fidelity are distinct evaluation dimensions within the same model interface. The conclusions are limited to the evaluated sites, horizons and driver ranges.

Future work will further improve the coordination between physical constraints and data-driven correction mechanisms and evaluate EWM under more complex environmental perturbations, across different regions and over longer forecast horizons. We will also extend EWM to Earth-surface systems in agriculture, hydrology and environmental ecology, with an emphasis on the coupled evolution of environmental drivers and system states across multiple timescales.

\FloatBarrier\bibliography{references}

\clearpage\section*{Supplementary Information}

\subsection*{S0 Experimental scope and reproducibility}
\setcounter{table}{-1}
\renewcommand{\thetable}{S\arabic{table}}

The four-site PV repeatability experiment uses 15 min sampling, 18 history and 13 future-drive channels, 40 epochs, stride eight and learning rate $10^{-3}$. For each horizon, all three seeds start from the same pretrained weights: the 4 h pretrained checkpoint for 4 h, and the 1 d checkpoint for both 12 h and 1 d. Wind v4e and v4h use 10 min sampling, a 24-step history and random initialization, with at most 60 epochs. Their seed-specific training trajectories and data splits are matched. EMA stopping uses smoothing parameter 0.6, a five-epoch burn-in and patience of 15; hybrid selection retains the lowest post-burn-in raw-validation-loss checkpoint visited.

Response-score aggregation follows \eqnref{eq:rho-aggregate}. Each run-level score is computed from the net per-unit response in \eqnref{eq:rho} and interpreted relative to its site- and horizon-specific analytic reference.

\par\vspace{4pt}\noindent\begin{minipage}{\linewidth}
\refstepcounter{table}{\small\textbf{Table \thetable. Experimental groups and comparison conditions}}\label{tab:s0}\par\vspace{4pt}
{\fontsize{8.5}{10.2}\selectfont\setlength{\tabcolsep}{3pt}
\begin{tabular}{>{\raggedright\arraybackslash}p{\dimexpr 0.25000\linewidth-2\tabcolsep\relax}>{\raggedright\arraybackslash}p{\dimexpr 0.25000\linewidth-2\tabcolsep\relax}>{\raggedright\arraybackslash}p{\dimexpr 0.25000\linewidth-2\tabcolsep\relax}>{\raggedright\arraybackslash}p{\dimexpr 0.25000\linewidth-2\tabcolsep\relax}}
\toprule
\textbf{Experimental group} & \textbf{Site / horizon} & \textbf{Repetitions and inputs} & \textbf{Comparison purpose} \\
\midrule
PV operation & Four sites; 4 h, 12 h, 1 d & Three seeds per cell; operational weather forecasts & Forecast performance; seed dispersion \\
Frozen PV audit & Six sites; 4 h, 12 h, 1 d & Three-seed median per cell & Response deficit; site conditions \\
PV decoder ablation & Six sites; average of 4 h and 1 d & One seed; shared encoder & Structural contribution to R\ensuremath{^{2}} and \ensuremath{\rho} \\
Matched wind controls & Kelmarsh; 4 h & Three seeds per model; shared split & Fit, pre-window timing and injection elasticity \\
\bottomrule\end{tabular}}\end{minipage}\par\vspace{4pt}

Initialization settings in this section refer to the four-site PV repeatability experiment and the matched wind comparison. The initialization of each six-site PV ablation is reported with its corresponding experiment. A seed labels a repeated training run under the same data split and model configuration.

No new training, inference or simulated observations are introduced. Error bars indicate sample SD across three seeds, not confidence intervals. The PV response tables report the net per-unit diagnostic defined in \eqnref{eq:rho}; numerical inversion errors are not used as evidence of accuracy.

\subsection*{S1 Four-site PV operational summaries}

\par\vspace{4pt}\noindent\begin{minipage}{\linewidth}
\refstepcounter{table}{\small\textbf{Table \thetable. Three-seed means and dispersion}}\label{tab:s1}\par\vspace{4pt}
{\fontsize{8.5}{10.2}\selectfont\setlength{\tabcolsep}{3pt}
\begin{tabular}{>{\raggedright\arraybackslash}p{\dimexpr 0.13648\linewidth-2\tabcolsep\relax}>{\raggedright\arraybackslash}p{\dimexpr 0.09095\linewidth-2\tabcolsep\relax}>{\raggedright\arraybackslash}p{\dimexpr 0.18191\linewidth-2\tabcolsep\relax}>{\raggedright\arraybackslash}p{\dimexpr 0.18191\linewidth-2\tabcolsep\relax}>{\raggedright\arraybackslash}p{\dimexpr 0.18191\linewidth-2\tabcolsep\relax}>{\raggedright\arraybackslash}p{\dimexpr 0.11372\linewidth-2\tabcolsep\relax}>{\raggedright\arraybackslash}p{\dimexpr 0.11312\linewidth-2\tabcolsep\relax}}
\toprule
\textbf{Site} & \textbf{Horizon} & \textbf{R\ensuremath{^{2}} \ensuremath{\pm} SD} & \textbf{MAE* \ensuremath{\pm} SD} & \textbf{MSE* \ensuremath{\pm} SD} & \textbf{n\_train} & \textbf{n\_test} \\
\midrule
S-1 & 4h & 0.8980 \ensuremath{\pm} 0.0020 & 0.2339 \ensuremath{\pm} 0.0046 & 0.2353 \ensuremath{\pm} 0.0046 & 2886 & 361 \\
S-1 & 12h & 0.8722 \ensuremath{\pm} 0.0127 & 0.2828 \ensuremath{\pm} 0.0218 & 0.2962 \ensuremath{\pm} 0.0296 & 2882 & 357 \\
S-1 & 1d & 0.8755 \ensuremath{\pm} 0.0043 & 0.2739 \ensuremath{\pm} 0.0016 & 0.2900 \ensuremath{\pm} 0.0100 & 2876 & 351 \\
S-2 & 4h & 0.9095 \ensuremath{\pm} 0.0018 & 0.8368 \ensuremath{\pm} 0.0132 & 3.2085 \ensuremath{\pm} 0.0639 & 3327 & 416 \\
S-2 & 12h & 0.9023 \ensuremath{\pm} 0.0078 & 0.8994 \ensuremath{\pm} 0.0774 & 3.4711 \ensuremath{\pm} 0.2780 & 3323 & 412 \\
S-2 & 1d & 0.9012 \ensuremath{\pm} 0.0037 & 0.9147 \ensuremath{\pm} 0.0358 & 3.5293 \ensuremath{\pm} 0.1314 & 3317 & 406 \\
S-3 & 4h & 0.8806 \ensuremath{\pm} 0.0007 & 0.7946 \ensuremath{\pm} 0.0013 & 3.0610 \ensuremath{\pm} 0.0179 & 3327 & 416 \\
S-3 & 12h & 0.8604 \ensuremath{\pm} 0.0009 & 0.8742 \ensuremath{\pm} 0.0108 & 3.5724 \ensuremath{\pm} 0.0231 & 3323 & 412 \\
S-3 & 1d & 0.8578 \ensuremath{\pm} 0.0048 & 0.8792 \ensuremath{\pm} 0.0209 & 3.6296 \ensuremath{\pm} 0.1207 & 3317 & 406 \\
S-4 & 4h & 0.9056 \ensuremath{\pm} 0.0006 & 0.7258 \ensuremath{\pm} 0.0026 & 2.5015 \ensuremath{\pm} 0.0170 & 3327 & 416 \\
S-4 & 12h & 0.8942 \ensuremath{\pm} 0.0052 & 0.8377 \ensuremath{\pm} 0.0244 & 2.8058 \ensuremath{\pm} 0.1375 & 3323 & 412 \\
S-4 & 1d & 0.8948 \ensuremath{\pm} 0.0040 & 0.8429 \ensuremath{\pm} 0.0332 & 2.7919 \ensuremath{\pm} 0.1070 & 3317 & 406 \\
\bottomrule\end{tabular}}\end{minipage}\par\vspace{4pt}

MAE and MSE retain their original numerical scales; units and inverse normalization require verification. The n\_train and n\_test counts are reused for each seed and are not multiplied by the number of seeds.

\subsection*{S2 PV decoder and six-site response audits}

\par\vspace{4pt}\noindent\begin{minipage}{\linewidth}
\refstepcounter{table}{\small\textbf{Table \thetable. Single-seed decoder ablation}}\label{tab:s2}\par\vspace{4pt}
{\fontsize{8.5}{10.2}\selectfont\setlength{\tabcolsep}{3pt}
\begin{tabular}{>{\raggedright\arraybackslash}p{\dimexpr 0.20000\linewidth-2\tabcolsep\relax}>{\raggedright\arraybackslash}p{\dimexpr 0.20000\linewidth-2\tabcolsep\relax}>{\raggedright\arraybackslash}p{\dimexpr 0.20000\linewidth-2\tabcolsep\relax}>{\raggedright\arraybackslash}p{\dimexpr 0.20000\linewidth-2\tabcolsep\relax}>{\raggedright\arraybackslash}p{\dimexpr 0.20000\linewidth-2\tabcolsep\relax}}
\toprule
\textbf{Site} & \textbf{MLP R\ensuremath{^{2}}} & \textbf{Physical R\ensuremath{^{2}}} & \textbf{MLP \ensuremath{\rho}} & \textbf{Physical \ensuremath{\rho}} \\
\midrule
S-1 & 0.818 & 0.853 & 0.174 & 0.940 \\
S-2 & 0.905 & 0.867 & 0.184 & 0.922 \\
S-3 & 0.856 & 0.815 & 0.198 & 0.871 \\
S-4 & 0.891 & 0.861 & 0.173 & 0.879 \\
P-CO-KERSEY & 0.852 & 0.792 & 0.117 & 0.945 \\
P-CA-ARBUCKLE & 0.831 & 0.762 & 0.238 & 0.974 \\
mean & 0.859 & 0.825 & 0.180 & 0.922 \\
\bottomrule\end{tabular}}\end{minipage}\par\vspace{4pt}

Values average the 4 h and 1 d results. The mean row retains the published rounded values rather than newly calculated exact means.

\subsection*{S3 Frozen-configuration PV audit}

\par\vspace{4pt}\noindent\begin{minipage}{\linewidth}
\refstepcounter{table}{\small\textbf{Table \thetable. Six sites and eighteen task cells}}\label{tab:s3}\par\vspace{4pt}
{\fontsize{8.5}{10.2}\selectfont\setlength{\tabcolsep}{3pt}
\begin{tabular}{>{\raggedright\arraybackslash}p{\dimexpr 0.16667\linewidth-2\tabcolsep\relax}>{\raggedright\arraybackslash}p{\dimexpr 0.16667\linewidth-2\tabcolsep\relax}>{\raggedright\arraybackslash}p{\dimexpr 0.16667\linewidth-2\tabcolsep\relax}>{\raggedright\arraybackslash}p{\dimexpr 0.16667\linewidth-2\tabcolsep\relax}>{\raggedright\arraybackslash}p{\dimexpr 0.16667\linewidth-2\tabcolsep\relax}>{\raggedright\arraybackslash}p{\dimexpr 0.16667\linewidth-2\tabcolsep\relax}}
\toprule
\textbf{Site} & \textbf{Arm} & \textbf{Horizon} & \textbf{R\ensuremath{^{2}}} & \textbf{\ensuremath{\rho}} & \textbf{\ensuremath{\Delta}\ensuremath{\rho}} \\
\midrule
S-1 & nodeb & 4h & 0.853 & 0.926 & -0.0001 \\
S-1 & nodeb & 12h & 0.853 & 0.945 & +0.0004 \\
S-1 & nodeb & 1d & 0.854 & 0.957 & +0.0007 \\
S-2 & nodeb & 4h & 0.889 & 0.927 & -0.0039 \\
S-2 & nodeb & 12h & 0.861 & 0.951 & +0.0033 \\
S-2 & nodeb & 1d & 0.862 & 0.961 & +0.0023 \\
S-3 & cfh & 4h & 0.817 & 0.851 & -0.0711 \\
S-3 & cfh & 12h & 0.807 & 0.887 & -0.0579 \\
S-3 & cfh & 1d & 0.812 & 0.892 & -0.0649 \\
S-4 & cfh & 4h & 0.862 & 0.864 & -0.0595 \\
S-4 & cfh & 12h & 0.865 & 0.883 & -0.0619 \\
S-4 & cfh & 1d & 0.860 & 0.894 & -0.0644 \\
P-CO-KERSEY & cfh & 4h & 0.791 & 0.900 & -0.0541 \\
P-CO-KERSEY & cfh & 12h & 0.789 & 0.953 & -0.0213 \\
P-CO-KERSEY & cfh & 1d & 0.793 & 0.975 & -0.0039 \\
P-CA-ARBUCKLE & nodeb & 4h & 0.758 & 0.982 & +0.0051 \\
P-CA-ARBUCKLE & nodeb & 12h & 0.755 & 0.982 & +0.0027 \\
P-CA-ARBUCKLE & nodeb & 1d & 0.761 & 0.968 & -0.0040 \\
\bottomrule\end{tabular}}\end{minipage}\par\vspace{4pt}

Each cell reports the median over three seeds. The response scores and signed differences use the net per-unit definition in \eqnref{eq:rho} and are interpreted relative to the analytic expectation. The source threshold $\Delta\rho\geq-0.03$ is applied to the median; seed-level values remain available for assessing run-to-run consistency. This table is distinct from the single-seed, two-horizon averages in \tabref{tab:s2} and the seed-0, 1 d diagnostic in \tabref{tab:s6}.

\subsection*{S4 PV inversion}

\par\vspace{4pt}\noindent\begin{minipage}{\linewidth}
\refstepcounter{table}{\small\textbf{Table \thetable. PV target-inversion specification}}\label{tab:s4}\par\vspace{4pt}
{\fontsize{8.5}{10.2}\selectfont\setlength{\tabcolsep}{3pt}
\begin{tabular}{>{\raggedright\arraybackslash}p{\dimexpr 0.25\linewidth-2\tabcolsep\relax}>{\raggedright\arraybackslash}p{\dimexpr 0.75\linewidth-2\tabcolsep\relax}}
\toprule
\textbf{Item} & \textbf{Definition} \\
\midrule
Queried drive & Future irradiance, scaled by $s$ \\
Target & $E(s^*)/E(1)=r^*$, with valid non-zero $E(1)$ \\
Matched conditions & Initial state, model weights, unqueried drivers and integration window \\
Recorded scope & 12 h and 1 d; operating target ratios $r^*\in[0.9,1.1]$ \\
Back-substitution & Apply the solved scale to the same forward model and compare the resulting energy ratio with the target \\
\bottomrule\end{tabular}}\end{minipage}\par\vspace{4pt}

This table defines the inversion query and its matched conditions. Quantitative accuracy assessment requires explicit solver settings and error normalization; the archived results cover the stated 12 h and 1 d conditions.

\subsection*{S5 Seed-level wind data}

\par\vspace{4pt}\noindent\begin{minipage}{\linewidth}
\refstepcounter{table}{\small\textbf{Table \thetable. Kelmarsh 4 h forecasting accuracy}}\label{tab:s5}\par\vspace{4pt}
{\fontsize{8.5}{10.2}\selectfont\setlength{\tabcolsep}{3pt}
\begin{tabular}{>{\raggedright\arraybackslash}p{\dimexpr 0.16667\linewidth-2\tabcolsep\relax}>{\raggedright\arraybackslash}p{\dimexpr 0.16667\linewidth-2\tabcolsep\relax}>{\raggedright\arraybackslash}p{\dimexpr 0.16667\linewidth-2\tabcolsep\relax}>{\raggedright\arraybackslash}p{\dimexpr 0.16667\linewidth-2\tabcolsep\relax}>{\raggedright\arraybackslash}p{\dimexpr 0.16667\linewidth-2\tabcolsep\relax}>{\raggedright\arraybackslash}p{\dimexpr 0.16667\linewidth-2\tabcolsep\relax}}
\toprule
\textbf{Model} & \textbf{Seed 0} & \textbf{Seed 1} & \textbf{Seed 2} & \textbf{Mean} & \textbf{Sample SD} \\
\midrule
v11 & 0.8070892 & 0.8077705 & 0.8095811 & 0.8081469 & 0.0012879 \\
m2sc & 0.8144378 & 0.8154114 & 0.8176929 & 0.8158474 & 0.0016708 \\
v4e & 0.7391756 & 0.6776420 & 0.7320113 & 0.7162763 & 0.0336495 \\
v4h & 0.7391756 & 0.7187808 & 0.7461483 & 0.7347016 & 0.0142217 \\
v4n & 0.8030000 & 0.7930000 & 0.7580000 & 0.7846667 & 0.0236291 \\
\bottomrule\end{tabular}}\end{minipage}\par\vspace{4pt}

Only three-decimal manuscript values are available for v4n; trailing zeros do not add precision. Other models use the JSON records.

\subsection*{S6 PV residual-freezing diagnostic}

\par\vspace{4pt}\noindent\begin{minipage}{\linewidth}
\refstepcounter{table}{\small\textbf{Table \thetable. Residual-freezing diagnostic under GHI \ensuremath{\times} 1.2 (seed 0, 1 d)}}\label{tab:s6}\par\vspace{4pt}
{\fontsize{8.5}{10.2}\selectfont\setlength{\tabcolsep}{3pt}
\begin{tabular}{>{\raggedright\arraybackslash}p{\dimexpr 0.20000\linewidth-2\tabcolsep\relax}>{\raggedright\arraybackslash}p{\dimexpr 0.20000\linewidth-2\tabcolsep\relax}>{\raggedright\arraybackslash}p{\dimexpr 0.20000\linewidth-2\tabcolsep\relax}>{\raggedright\arraybackslash}p{\dimexpr 0.20000\linewidth-2\tabcolsep\relax}>{\raggedright\arraybackslash}p{\dimexpr 0.20000\linewidth-2\tabcolsep\relax}}
\toprule
\textbf{Site} & \textbf{Analytic expectation} & \textbf{Fixed \ensuremath{\eta}} & \textbf{Full model} & \textbf{\ensuremath{\Delta}\ensuremath{\rho}} \\
\midrule
S-1 & 0.9558 & 0.9558 & 0.9551 & -0.0007 \\
S-2 & 0.9585 & 0.9585 & 0.9135 & -0.0450 \\
S-3 & 0.9569 & 0.9569 & 0.8951 & -0.0619 \\
S-4 & 0.9579 & 0.9579 & 0.8986 & -0.0593 \\
P-CO-KERSEY & 0.9795 & 0.9795 & 0.9827 & 0.0033 \\
P-CA-ARBUCKLE & 0.9718 & 0.9718 & 0.9678 & -0.0040 \\
\bottomrule\end{tabular}}\end{minipage}\par\vspace{4pt}

\figfullref{fig:interventions}A plots absolute values of the source signed \ensuremath{\Delta}\ensuremath{\rho} in this table; lower mismatch indicates closer agreement with the analytic expectation. Holding \ensuremath{\eta} fixed retains the pre-intervention site correction, not unchanged power. The site--horizon coverage discussed in \secreftext{sec:inversion}{Section~4.4.3} uses the three-run medians in \tabref{tab:s3}. \figfullref{fig:interventions}B reports descriptive counts, and \figref{fig:interventions}C uses 64-window extrema, not confidence intervals.

\end{document}